\pdfoutput=1
\def\ICLRArxivVersion{1}
\def\ICLRIncludeAppendix{1}
\documentclass{article} 
\usepackage{iclr2027_conference,times}

\usepackage{amsmath,amsfonts,bm}

\def\eqref#1{equation~\ref{#1}}

\def\1{\bm{1}}

\DeclareMathAlphabet{\mathsfit}{\encodingdefault}{\sfdefault}{m}{sl}
\SetMathAlphabet{\mathsfit}{bold}{\encodingdefault}{\sfdefault}{bx}{n}

\usepackage{algorithm}
\usepackage{algpseudocode}
\ifdefined\ICLRIncludeAppendix
\else
\usepackage{xr-hyper}
\fi
\usepackage[hidelinks]{hyperref}
\ifdefined\ICLRIncludeAppendix
\else
\fi
\usepackage{url}
\usepackage{graphicx}
\usepackage{booktabs}
\usepackage{array}
\usepackage{longtable}
\usepackage{float}
\usepackage{multirow}
\usepackage{xcolor}
\usepackage[section]{placeins}
\newcolumntype{P}[1]{>{\raggedright\arraybackslash}p{#1}}
\title{Beyond Correctness: Evaluating Semantic Knowledge in Cross-Table Transfer}
\ifdefined\ICLRArxivVersion
\author{\normalfont
\textbf{Seokyong Sheem}\thanks{Equal contribution.} \quad \textbf{Hochang Lee}\footnotemark[1] \quad \textbf{Suyeong Lee} \quad \textbf{Daekyum Kim}\thanks{Corresponding author.} \\
Korea University \\
\texttt{\{sheemsy, cvcvcv99, tsw1615, daekyum\}@korea.ac.kr}
}
\else
\author{Anonymous Authors}
\fi

\ifdefined\ICLRArxivVersion
\iclrfinalcopy
\fi

\newcommand{\supp}[1]{\textnormal{Supp.~#1}}
\usepackage{enumitem}

\begin{document}

\maketitle
\ifdefined\ICLRArxivVersion
\lhead{}
\fi
\begin{abstract}

Semantic knowledge is increasingly used to bridge heterogeneous schemas in tabular learning, but how much does that knowledge actually improve prediction?
Studies in tabular learning commonly answer this question through semantic ablations that modify or suppress the supplied semantic knowledge.
We show that these ablations can lead to misleading conclusions about predictive benefit: poor performance under altered semantics may be taken as evidence that the intended knowledge is beneficial.
Across real and controlled experiments, altering semantic content can produce large performance differences even when the model gains little predictive benefit from having that semantic knowledge in the first place.
To separate these effects, we distinguish two quantities: content sensitivity and predictive utility. 
Content sensitivity measures the change in performance when semantic content is altered, whereas predictive utility measures the benefit of the intended semantic knowledge relative to a suitable reference without that knowledge.
This distinction motivates an evaluation framework in which the control is chosen according to the question being asked: altered controls assess sensitivity to semantic content, whereas claims that semantic knowledge improves prediction require a suitable reference.
Even then, predictive utility is not fixed; it varies across suitable references and decreases when the reference can more easily recover the tested knowledge from other inputs or labeled examples.
In a bounded audit of 25 semantic-ablation comparisons across nine studies, only one of 18 explicit predictive-utility claims is paired with a control that clearly isolates the tested semantic contribution.
Together, these findings motivate a simple evaluation principle: semantic-ablation controls should be chosen and interpreted according to the question they are intended to answer. 

Code and reproducibility artifacts are available at \url{https://github.com/mintlabkorea/beyond-correctness}.

\end{abstract}

\section{Introduction}

While tabular datasets record data in a structured format, they often organize similar information under different schemas. 
Across datasets, corresponding variables may differ in their names, coding conventions, units, scales, or measurement protocols, even when they represent the same underlying concept. 
Predictive models typically rely on schema-specific input representations, making direct transfer difficult when the source and target schemas differ. 

Recent approaches to cross-table transfer and pretrained tabular learning increasingly leverage semantic knowledge to interpret heterogeneous schemas and enable transfer across datasets \citep{wang2022transtab,kim2024carte,kim2025tarte,arazi2025tabstar}.
Such knowledge can come from feature names, dataset documentation, codebooks, knowledge graphs, or domain rules, and can help interpret recorded values, identify corresponding variables, and introduce structured relationships useful for prediction \citep{pang2015biobankconnect,li2025harmonization,ruiz2023plato,han2024featllm}.
As such knowledge becomes increasingly integrated into tabular learning, it becomes important to determine how much the supplied knowledge contributes to predictive performance. 

Studies in tabular learning commonly address this question through semantic ablations that modify or suppress the supplied semantic knowledge \citep{dinh2022lift,hegselmann2023tabllm}.
For example, feature names or other semantic inputs may be shuffled, permuted, anonymized, or omitted, and the resulting performance is compared with that obtained using the intended semantic knowledge \citep{dinh2022lift,hegselmann2023tabllm}.
However, differences in performance between intended and altered conditions do not necessarily mean that the intended knowledge improves prediction \citep{hewitt2019designing,elazar2021amnesic,sturmfels2020visualizing}.
Such differences can arise when changing the semantic content reduces performance in the altered condition.
The resulting performance gap therefore does not isolate the benefit of the intended knowledge, as it also captures degradation from the altered condition.
In such cases, the ablation may be misinterpreted as showing that the intended semantic knowledge improves prediction relative to a condition without that knowledge.

To separate these effects, we distinguish two evaluation quantities: content sensitivity and predictive utility. 
Content sensitivity measures how performance changes when the supplied semantic content is altered, whereas predictive utility measures the benefit of the intended semantic knowledge relative to a suitable reference in which the tested component is unavailable. 
The two quantities answer different questions: content sensitivity asks whether prediction depends on which semantic content is supplied, whereas predictive utility asks whether having the tested knowledge improves prediction in the first place.
Evaluating predictive utility therefore requires a suitable reference that satisfies three criteria: (1) Removal: remove the tested semantic component; (2) Preservation: preserve all other model inputs and how they are presented; and (3) Setup: keep the model and learning procedure unchanged.
This leads to a claim-matched evaluation principle: the control should be chosen according to the question the comparison is intended to answer.

\begin{figure}[t]
\centering
\includegraphics[width=\linewidth]{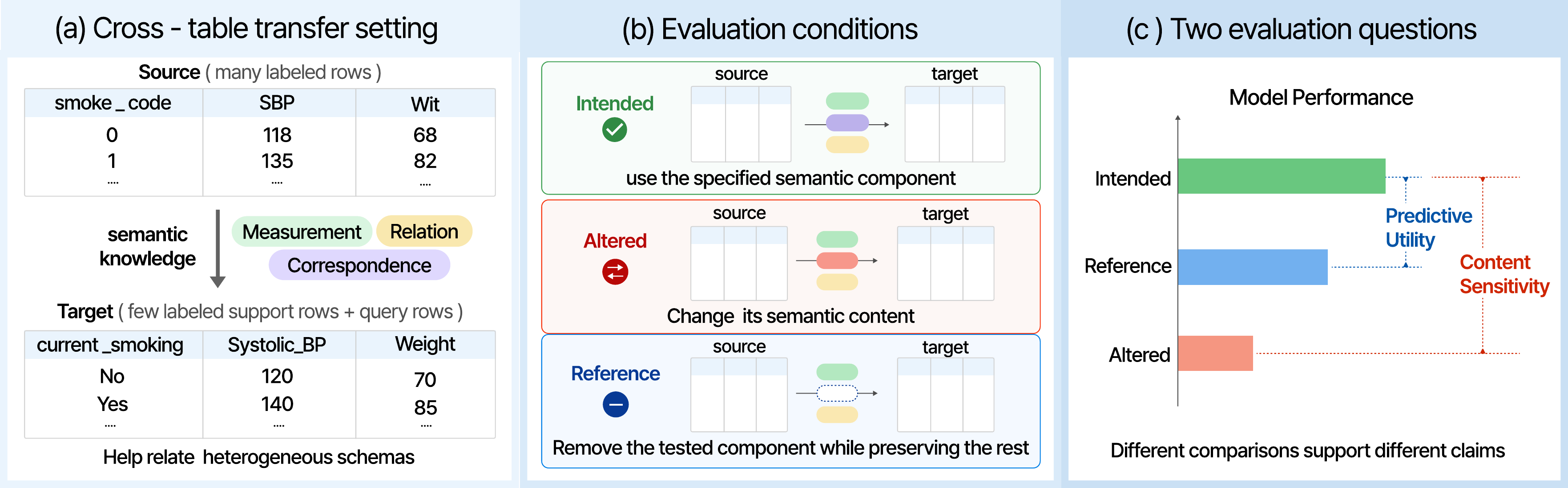}
\caption{
\textbf{Overview of the evaluation framework.}
(a) Semantic knowledge relates heterogeneous source and target schemas.
(b) The intended condition uses the specified content, the altered condition changes it, and the reference removes the tested component while preserving the remaining setup.
(c) Intended--altered measures \emph{content sensitivity}, whereas intended--reference measures \emph{predictive utility}; the two comparisons support different claims.
}

\label{fig:overview}
\vspace{-10pt}

\end{figure}

Our experiments show that these distinctions matter in practice. 
Across an existing semantic tabular model, real cross-table transfer tasks, and controlled experiments, altering semantic content can produce large performance differences even when the intended knowledge provides little predictive benefit over a suitable reference. 
In several settings, much of the intended–altered difference is explained by degradation under the altered condition rather than improvement attributable to the intended knowledge. 
The effect is not resolved simply by choosing a suitable reference: predictive utility itself varies across suitable references and decreases when the reference can more easily recover the tested information from other observed inputs or labeled target examples. 
Thus, the measured benefit depends not only on the knowledge being evaluated, but also on what the comparison condition can recover without it.

The same issue is visible in published practice. 
In a bounded audit of 25 semantic-ablation comparisons across nine studies, only one of 18 explicit predictive-utility claims is paired with a control satisfying all three requirements for a suitable reference; 11 fail at least one requirement and six are indeterminate from the published description. 
This does not imply that the corresponding semantic knowledge is unhelpful.
Rather, it means that the reported comparison does not by itself isolate the claimed predictive utility. 
Together, these findings motivate evaluation designs in which semantic-ablation controls are selected and interpreted according to the question they are intended to answer.  

\noindent\textbf{Contributions.}
\begin{itemize}[leftmargin=1.2em, labelsep=0.4em,
                itemsep=1pt, topsep=1pt, parsep=0pt, partopsep=0pt]

\item \textbf{Conceptual.} 
We show that semantic-ablation effects can conflate predictive benefit with degradation induced by the control, and distinguish content sensitivity from predictive utility to separate these effects. 
\item \textbf{Methodological.} 
We propose claim-matched evaluation with criteria for constructing suitable references and for assessing sensitivity to the choice of reference.  
\item \textbf{Empirical.} 
Through a reanalysis of an existing semantic tabular model, real cross-table transfer tasks, controlled experiments, and published ablations, we identify when and why semantic-ablation effects diverge from predictive utility and how measured utility changes with the information available to the reference. 

\end{itemize}

\section{Related Work}

\subsection{Semantic Knowledge in Cross-Table Transfer}

Cross-table methods such as TransTab and CARTE use semantic knowledge to relate heterogeneous schemas: TransTab incorporates feature descriptions, while CARTE uses open-vocabulary representations \citep{wang2022transtab,kim2024carte}. Related semantic tabular methods, such as TabLLM, serialize feature names and values as text \citep{hegselmann2023tabllm}.
More broadly, semantic or structured information in tabular learning can come from dataset documentation, schema-matching systems, knowledge graphs, language-model-generated features and rules, or task-derived behavioral constraints \citep{pang2015biobankconnect,li2025harmonization,ruiz2023plato,han2024featllm,kim2025kinematics}.
These works primarily study how such information is obtained, represented, or used, leaving open whether a particular use improves prediction.

\subsection{Evaluating the Contribution of Semantic Knowledge}

Across semantic tabular learning, ablations often compare standard semantic inputs with altered alternatives, such as shuffled or permuted inputs, or with omission-based controls, as in LIFT and TabLLM \citep{dinh2022lift,hegselmann2023tabllm}.
Because these controls can change different aspects of the input, a performance gap may reflect more than the contribution of the tested semantic knowledge.
Related work on probing, information removal, and attribution similarly shows that conclusions depend on what a control changes, what it keeps fixed, and the chosen reference point \citep{hewitt2019designing,fisher2019all,elazar2021amnesic,sturmfels2020visualizing}.
Separately, transfer-learning studies show that the value of transferred information can diminish as labeled target support increases \citep{he2019rethinking,tripuraneni2020theory,zoph2020rethinking,nguyen2020leep}.
We therefore evaluate semantic interventions using claim-matched controls, distinguishing sensitivity to semantic content from predictive utility relative to a suitable reference and examining how utility varies across suitable references.

\section{Evaluating the Contribution of Semantic Knowledge}
\label{sec:method}

\subsection{Evaluation Design}
\label{sec:evaluation_design}

We evaluate each semantic component under matched conditions, holding the evaluation data, model, training procedure, and all other components fixed.
Our experiments use a labeled source table and a target table, with labeled target rows for adaptation and the remaining rows for evaluation.
In the TabLLM reanalysis, semantic knowledge is instead varied within the model input.

We study three recurring roles of semantic knowledge.
\emph{Measurement} knowledge interprets coding, polarity, scales, units, and missing-value meanings;
\emph{correspondence} knowledge identifies compatible variables across schemas;
and \emph{relation} knowledge specifies structure among variables, such as derived values or directional constraints.
These roles organize the test cases and are not treated as a full taxonomy.

For each component, the \emph{intended condition} uses its prespecified form.
We use \emph{control} as a general term for a condition against which the intended condition is compared.
An \emph{altered control} changes the tested semantic content while preserving how it is used in prediction, whereas a \emph{removal control} omits that content.
A removal control serves as a \emph{suitable reference} only if it satisfies the criteria in Section~\ref{sec:reference}.
Let $Q_c$ denote performance under condition $c$.
All conditions use the same evaluation rows and procedure, and higher is better for every metric.
We define

\begin{align}
\text{Predictive utility}
    &= Q_{\mathrm{intended}} - Q_{\mathrm{reference}},\\
\text{Content sensitivity}
    &= Q_{\mathrm{intended}} - Q_{\mathrm{altered}}.
\end{align}

\begin{table}[h]
\caption{Condition construction for measurement, correspondence, and relation knowledge.}
\label{tab:knowledge_conditions}
\centering
\fontsize{8pt}{9pt}\selectfont
\setlength{\tabcolsep}{3pt}
\renewcommand{\arraystretch}{0.92}

\begin{tabular}{@{}P{0.16\linewidth}P{0.25\linewidth}P{0.29\linewidth}P{0.25\linewidth}@{}}
\toprule
Knowledge & Intended & Altered & Reference \\
\midrule

Measurement &
Specified coding or numerical conversion &
Altered coding or numerical conversion &
Raw source values \\

Correspondence &
Specified source--target pairing &
Permuted source--target pairing &
No shared pairing; separate source and target columns \\

Relation &
Specified relation and direction &
Reversed direction or random relation &
No added value or constraint \\

\bottomrule
\end{tabular}
\end{table}

For correspondence, the reference keeps the same number of columns but places source and target values in separate columns rather than a shared matched column.
In the controlled benchmark, known schema transformations define the intended measurement and correspondence knowledge, while the synthetic outcome mechanism defines the intended relations.
In real-data experiments, intended knowledge is derived from documentation or domain sources before evaluation and is not assumed to be ground-truth correct.

\subsection{Constructing a Reference}
\label{sec:reference}

\begin{figure}[h]
\centering
\includegraphics[width=0.9\linewidth]{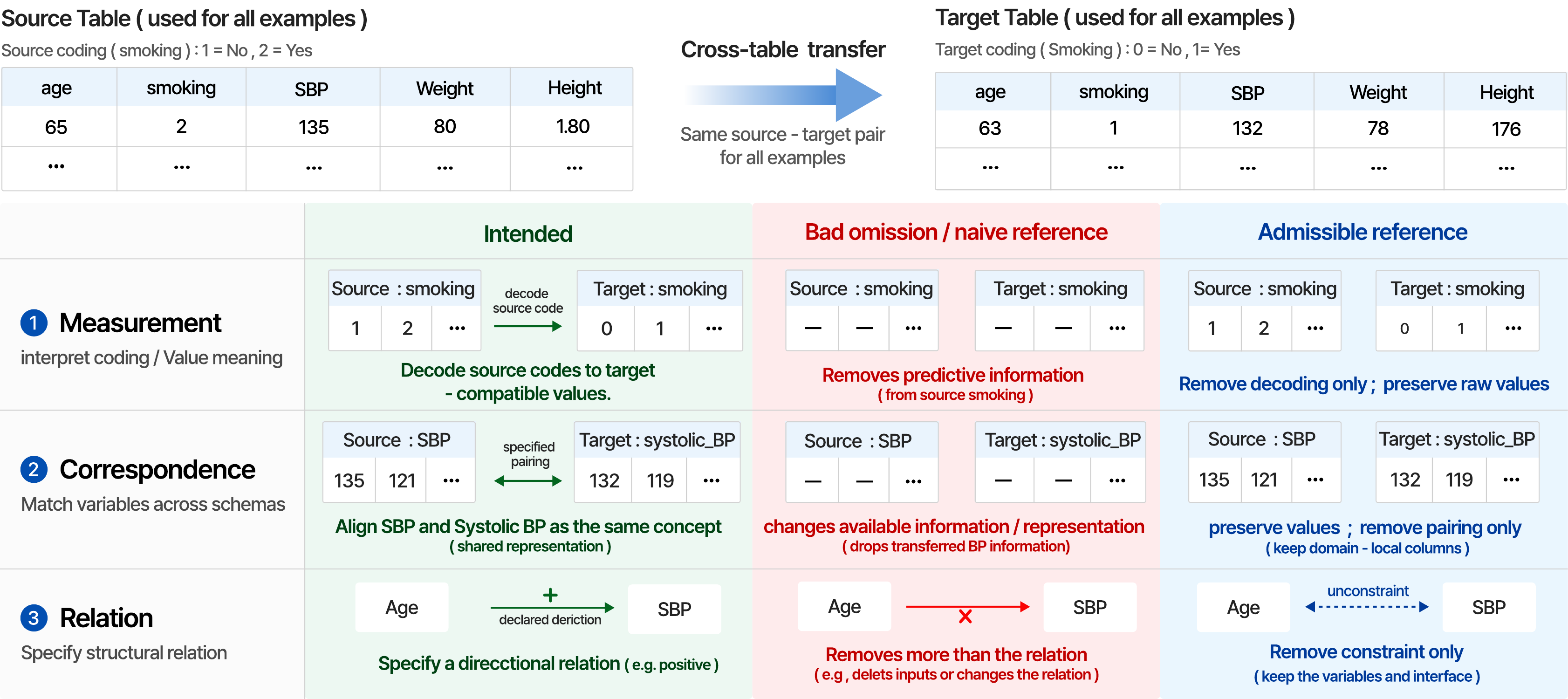}
\caption{
\textbf{Suitable and unsuitable references for the three roles of semantic knowledge. }
Suitable references remove only the tested component while preserving the remaining predictive information and setup. 
Unsuitable controls either remove additional predictive information or alter, rather than remove, the tested component.
}
\label{fig:reference_construction}
\end{figure}

A suitable reference must remove the evaluated semantic component while preserving the rest of the comparison.
Simply removing the component may also change the input; for example, removing feature names can turn a key--value input into a values-only input (Figure~\ref{fig:reference_construction}).

Removing the evaluated component yields a candidate reference, which is suitable only if it satisfies three criteria:
(i) \emph{Removal}: it removes the evaluated semantic component;
(ii) \emph{Preservation}: preserves all other information available to the model and how that information is presented, including feature identity, input width, and format;
(iii) \emph{Setup}: leaves the model, learning procedure, and other modeling choices unchanged.

For text inputs, changing the tested text can change tokens; this alone does not fail Preservation. 
For feature-name tests, Removal means that the original feature-name meaning is removed; we do not assume that every anonymous label has exactly the same effect on the model.
In the published-ablation audit, we check Preservation by asking whether the other predictive values, feature identity, input width, and the rest of the input format are preserved, and check Setup by asking whether the model and learning procedure are unchanged \supp{\S\ref{sec:published_audit}}.

Controls that fail these criteria may support broader comparisons but do not isolate predictive utility.
Predictive utility is defined relative to a particular suitable reference, so different suitable references need not give the same value.
If multiple references satisfy these criteria, we report reference sensitivity; if none does, predictive utility is not separately evaluable.

For a specified suitable reference, predictive utility is positive or negative when its 95\% confidence interval lies entirely above or below zero; otherwise it is unresolved.
Thus, \emph{not separately evaluable} means that no suitable reference exists, whereas \emph{unresolved} means that a suitable reference exists but the effect direction is not resolved.
Resolution diagnostics are reported in \supp{\S\ref{sec:app-controlled-boundaries}}.

In practice, claim-matched evaluation states the claim, selects a matching altered or removal control, and, for predictive utility, verifies Removal, Preservation, and Setup before reporting the contrast.

\subsection{Constructing Semantic Knowledge}
\label{sec:knowledge_construction}

To evaluate the contribution of each role of semantic knowledge separately, we construct measurement, correspondence, and relation knowledge.
Target labels and predictive performance are not used to create, select, or revise these specifications.

Measurement knowledge is obtained from dataset documentation, including codebooks, value labels, questionnaire wording, units, and measurement protocols.
These sources determine coding, polarity, missing-value meanings, and documented unit conversions.
We do not infer interpretations from observed values when documentation is missing or contradictory.

For correspondence knowledge, GPT-5.6 Sol proposes groups of related source and target variables from names, descriptions, units, value definitions, and dataset documentation, without access to row values or target labels.
Groups may contain multiple variables from either schema and must cite the supplied documentation and pass the fixed validation checks described in the supplement.

For relation knowledge, candidates are constructed from dataset documentation and fixed before evaluation.
Each retained relation specifies the variables, an executable operation, and a direction when applicable.
After fixing the relation set, we review the literature only to document support; the review does not add, remove, or revise relations.
The supplement provides prompts, schemas, validation rules, and execution records for the LLM-based construction step (\supp{\S\ref{sec:app-extraction-documentation}}).

\section{Experiments and Results}
\label{sec:experiments}

\subsection{Experimental Settings}
\label{sec:exp_setup}

Our quantitative evaluation spans three settings summarized in Table~\ref{tab:experimental_settings}: the TabLLM reanalysis, real cross-table transfer, and the controlled benchmark.
Clinical experiments use NHANES (NH), KNHANES (KN), and HRS
\citep{johnson2013nhanes,kweon2014knhanes,sonnega2014hrs}.
Hydrological experiments use CAMELS-US hydrometeorological time series and
catchment attributes \citep{newman2014camelsdata,addor2017camelsattributes}
and CAMELS-GB v2 \citep{coxon2025camelsgbv2}.
Clinical tasks include binary questionnaire endpoints and continuous laboratory or anthropometric endpoints.
The TabLLM reanalysis examines a semantic ablation in an existing model, while the clinical, hydrological, and controlled experiments evaluate cross-table transfer directly.

\begin{table}[h]
\caption{Summary of the experimental settings.}
\label{tab:experimental_settings}
\centering
\fontsize{8pt}{9pt}\selectfont
\setlength{\tabcolsep}{3pt}

\begin{tabular}{@{}p{0.12\linewidth}p{0.45\linewidth}p{0.1\linewidth}p{0.22\linewidth}@{}}
\toprule
Setting & Data and task & Primary model & Metric \\
\midrule

TabLLM\newline reanalysis &
Nine original-study datasets; released-code reanalysis with the name-masked reference &
TabLLM\newline (released) &
AUROC \\

\midrule

Clinical &
NH--KN and pooled NH/KN--HRS; binary questionnaire and continuous examination endpoints &
XGB &
AUROC (binary); \newline negative SD-normalized RMSE (continuous) \\

\addlinespace

Hydrological &
CAMELS-US to CAMELS-GB v2; \newline 120 basins (CAMELS-120) &
XGB &
Negative log-RMSE \\

\midrule

Controlled\newline benchmark &
Eight-variable NHANES panel with observed values and missingness retained; constructed schemas and synthetic targets &
XGB &
Negative nMSE \\

\bottomrule
\end{tabular}

\vspace{2pt}

\begin{minipage}{\linewidth}
\fontsize{8pt}{9pt}\selectfont
\emph{Note.}
Higher is better for all metrics.
XGB is the primary cross-table learner; TransTab and CARTE are additionally evaluated in model-coverage analyses \supp{\S\ref{sec:app-learner-coverage}}.
For the controlled benchmark, $\mathrm{nMSE}=\mathrm{MSE}/\mathrm{Var}(y)$.
Full dataset, support, seed, learner, and inference settings are provided in \supp{\S\ref{sec:app-experimental-details}}.
\end{minipage}
\end{table}

Within each comparison, the model and fitting procedure are otherwise unchanged.

For the controlled benchmark, we use real NHANES feature values while varying schema mismatch and prediction-task structure independently.
Source and target schemas differ in coding polarity, numerical scale, missing-value coding, and column identifiers.
Synthetic outcomes follow additive, pairwise, or sparse functions, so the schema differences and outcome mechanisms are fully specified.

For the TabLLM reanalysis, we use the 11B T0/IA3 checkpoint with a name-masked reference that removes feature-name semantics while preserving feature identity and input format.
Our released-code compatibility reproduction puts 21 of 27 published means within $.015$ AUROC \supp{\S\ref{sec:app-tabllm-reproduction}}.

We use $n_{\mathrm{adapt}}$ for labeled TabLLM adaptation examples and $n_{\mathrm{target}}$ for labeled target examples in cross-table transfer.
Conditions share the same split and adaptation or target samples, and the same realization in the controlled benchmark, so all contrasts are paired.
Unless otherwise stated in the supplement, confidence intervals use 10,000 paired
resamples over endpoint--seed pairs for clinical experiments, basins for hydrology,
realizations for the controlled benchmark, and test rows for TabLLM.
Prespecified analyses form the primary results; post-hoc analyses are explicitly labeled and are accompanied by the corresponding broader analysis when available \supp{\S\ref{sec:app-experimental-details}}.
We use these settings to test three questions in sequence: whether content sensitivity reflects predictive utility (\ref{sec:gap_composition}), whether utility varies across suitable references (\ref{sec:utility_relative}), and how utility changes with the information available to the reference (\ref{sec:reference_recover}).

\subsection{Content Sensitivity Does Not Identify Predictive Utility}
\label{sec:gap_composition}

A large intended--altered gap does not necessarily measure predictive utility because it can also reflect degradation under the altered condition.
Content sensitivity can be decomposed as
\begin{equation}
\text{Content sensitivity} =
(\underbrace{ Q_{\mathrm{intended}}-Q_{\mathrm{reference}}}_{\text{predictive utility}})
+
(Q_{\mathrm{reference}}-Q_{\mathrm{altered}}).
\label{eq:decomposition}
\end{equation}
Equation~\ref{eq:decomposition} is an identity; empirically, we ask how large the reference--altered term is and whether it explains most of the observed intended--altered gap.
We apply this decomposition to TabLLM's feature-name ablation, real-data relation transfer, and the controlled benchmark.
In the relation experiments, switching from reversed to random directions changes only the altered condition, while the intended and reference conditions remain fixed.
Predictive utility therefore remains unchanged, allowing us to examine directly how the altered condition affects content sensitivity.
The controlled benchmark provides the analogous comparison when the intended relation is known exactly from the synthetic outcome design.
Table~\ref{tab:gap_utility} summarizes the decomposition.

\begin{table}[h]
\caption{Decomposition of content sensitivity across the evaluated settings.}
\label{tab:gap_utility}

\centering
\fontsize{8pt}{9pt}\selectfont
\setlength{\tabcolsep}{3pt}

\begin{tabular}{@{}llccc@{}}

\toprule
Setting & Altered condition & Content sensitivity & Predictive utility & Reference $-$ Altered \\
\midrule

\multicolumn{5}{@{}c}{\emph{TabLLM feature-name ablation}} \\
9-dataset mean
& \emph{List Permuted Names}
& $+.112$
& $+.088$
& $+.024$ \\

Credit-g
& \emph{List Permuted Names}
& $+.066$
& $-.003$
& $+.069$ \\

Bank
& \emph{List Permuted Names}
& $-.032$
& $+.089$
& $-.121$ \\

\midrule
\multicolumn{5}{@{}c}{\emph{Real cross-table transfer: relation knowledge}} \\
\multirow{2}{*}{NH--KN}
& Reversed & $+.183$ & $-.002$ & $+.185$ \\
& Random   & $-.001$ & $-.002$ & $+.001$ \\

\addlinespace[1pt]
\multirow{2}{*}{NH/KN--HRS}
& Reversed & $+.266$ & $+.002$ & $+.264$ \\
& Random   & $+.041$ & $+.002$ & $+.039$ \\

\addlinespace[1pt]
\multirow{2}{*}{CAMELS-120}
& Reversed & $+.144$ & $.000$ & $+.144$ \\
& Random   & $+.002$ & $.000$ & $+.002$ \\

\midrule
\multicolumn{5}{@{}c}{\emph{Controlled benchmark}} \\
\multirow{2}{*}{\shortstack[l]{Additive XGB\\($n_{\mathrm{target}}=32$)}}
& Reversed & $+.304$ & $+.024$ & $+.280$ \\
& Random   & $+.126$ & $+.024$ & $+.102$ \\

\bottomrule
\end{tabular}

\vspace{2pt}

\noindent
\begin{minipage}{\linewidth}
\fontsize{8pt}{9pt}\selectfont
\emph{Note.}
For TabLLM, predictive utility uses the name-masked reference.
Metrics are AUROC (TabLLM), negative SD-normalized RMSE (NH--KN and NH/KN--HRS), negative log-RMSE (CAMELS-120), and negative nMSE (controlled benchmark); higher is better.
Reference $-$ altered is the second term in Equation~\ref{eq:decomposition}; positive values mean that the altered condition performs worse than the reference.
Additional intervals and relation-control details are in \supp{Tables~\ref{tab:app-tabllm-reproduction} and~\ref{tab:app-relation-altered-construction}}.
\end{minipage}

\end{table}

TabLLM shows that content sensitivity can either overstate or understate predictive utility. 
On Credit-g, content sensitivity is $+.066$, whereas predictive utility is $-.003$, because the altered condition is $.069$ lower than the reference.
Bank shows the reverse case: the altered condition outperforms the reference by $.121$, so content sensitivity understates predictive utility ($-.032$ versus $+.089$).

The relation experiments show the same distinction.
Replacing reversed with random directions sharply reduces content sensitivity in all three real-data settings while leaving predictive utility unchanged.
The controlled benchmark shows the same pattern when the intended relation is known exactly from the synthetic outcome design.
For NH--KN with random directions, content sensitivity ($-.001$) nearly matches predictive utility ($-.002$) because the altered condition performs similarly to the reference.
For the reversed controls, the near-zero utility point estimates show that almost the entire intended--altered gap comes from degradation under the altered condition rather than improvement over the reference.
CAMELS-120 has a narrow 95\% CI around zero ($[-.001,+.001]$), whereas the clinical intervals remain wide.

The same separation also appears with TransTab. 
At $n_{\mathrm{target}}=32$, 74--80\% of the intended--altered gap is due to altered-condition degradation; at $n_{\mathrm{target}}=512$, this fraction rises to 97--100\%.
Additional CARTE results are more mixed, further indicating that the magnitude and direction of predictive utility also depend on the learner and how the semantic knowledge is supplied \supp{\S\ref{sec:app-learner-coverage}}.
Across these settings, the intended--altered gap depends on how the altered condition is constructed and need not reflect predictive utility.

\subsection{Predictive Utility Can Vary Across References}
\label{sec:utility_relative}

\begin{figure}[h]
\centering
\includegraphics[width=0.90\linewidth]{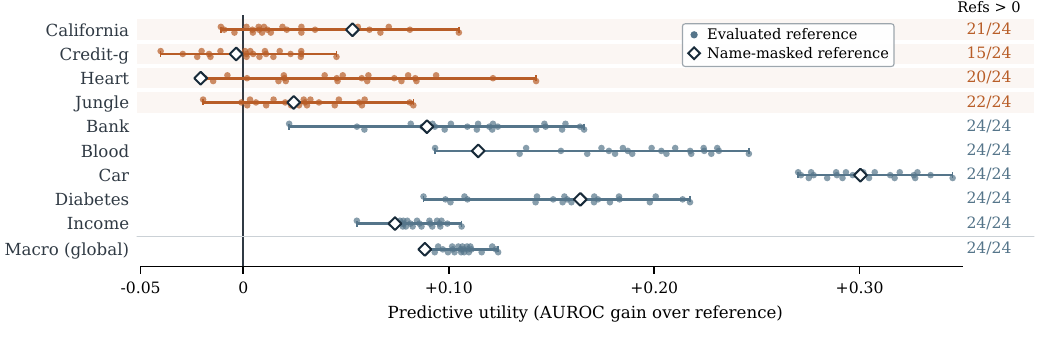}
\caption{
Reference sensitivity of TabLLM predictive utility.
Points show the 24 evaluated identifier assignments, diamonds mark the primary name-masked reference, and horizontal lines show the point-estimate range; orange rows indicate ranges that cross zero.
Counts at right show the number of positive estimates.
}
\label{fig:ref_sensitivity}
\end{figure}

We evaluate 24 name-masked references that remove the original feature-name meanings but assign generic identifiers differently; values and the remaining prompt structure are unchanged
The nine-dataset mean utility remains positive under all 24 references, but dataset-level point-estimate ranges cross zero for four of nine datasets (Figure~\ref{fig:ref_sensitivity}).

Across datasets, $79.2-99.7\%$ of the variation across references remains after accounting for evaluation noise with paired resampling over shared test rows.
Only for Jungle do simultaneous 95\% confidence bands confirm a sign change, with one band entirely below zero and another entirely above zero.
The other three crossings rest on point estimates only and remain unresolved.
Reference choice clearly affects utility magnitude, while evidence that it changes the resolved sign is much more limited.
Reference-dependent variation remains when we restrict the analysis to the four datasets whose published TabLLM results we reproduce most closely. 
Similar variation also appears with alternative generic identifier labels, although zero crossings are less frequent.
Satisfying the reference criteria allows predictive utility to be estimated, but different suitable references can still yield different values.
We recommend reporting reference sensitivity when multiple suitable references are available.

\subsection{Predictive Utility Depends on What the Reference Can Recover}
\label{sec:reference_recover}

We next test this dependence in two complementary ways: making the tested information harder to recover from observed inputs and easier to recover from labeled target examples.
As a real-data reference point, measurement knowledge has positive utility across 10 clinical endpoints ($+.197$, 95\% CI $[+.110,+.294]$).

\paragraph{Utility increases when information is harder to recover from observed inputs.}

In the controlled benchmark, we test how measurement utility changes as source--target measurement mismatch increases, making the required transformation harder for the reference to infer.
For relation knowledge, we instead remove constituent inputs, making the supplied relation value harder to recompute.
Within each comparison, the intended and reference conditions receive the same observed inputs \supp{\S\ref{sec:app-measurement-dose}, \S\ref{sec:app-relation-value-interface}}.
Across both interventions, utility increases as the tested information becomes harder to recover.
At the largest measurement mismatch and $n_{\mathrm{target}}=32$, XGB utility reaches $+.259$ to $+.260$ across the three target structures.
Relation utility similarly rises from $+.002$ to $+.242$ as the constituent inputs are removed.
This increase comes mainly from deterioration of the reference rather than improvement of the intended condition.
In the pairwise XGB setting, intended nMSE changes only from $.708$ to $.741$, while reference nMSE worsens from $.710$ to $.983$.

Removing constituent inputs changes the information available to the model. To test whether reconstructibility is related to utility, we add noise to the constituent inputs while keeping the number of input columns fixed.
At each nonzero noise level, lower reconstruction $R^2$ shows a modest association with higher utility (within-setting Spearman $\rho$ between $1-R^2$ and utility, averaged over eight settings at each dose: $+.165$ to $+.329$; \supp{\S\ref{sec:app-noisy-proxy}}).
Because this relationship cannot be isolated from other input changes in the real-data relation experiments, we treat it as a modest association observed in the controlled benchmark rather than a mechanism established in real-data transfer.

\begin{figure}[h]
\centering
\includegraphics[width=0.95\linewidth]{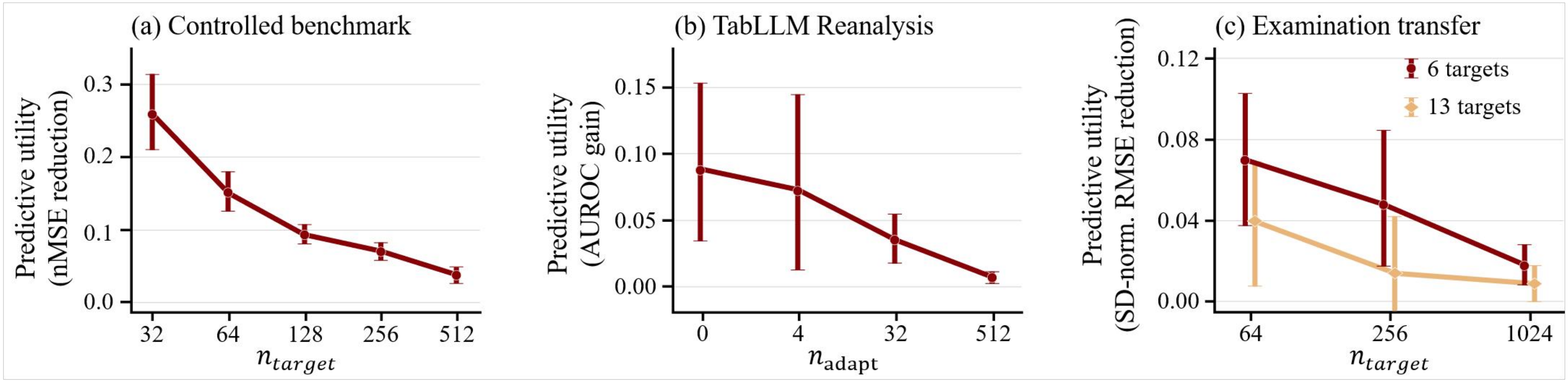}
\caption{
Predictive utility decreases as labeled target support increases.
\textbf{(a)} Controlled measurement-knowledge utility for pairwise XGB at the largest tested measurement mismatch.
\textbf{(b)} Mean TabLLM feature-name utility across nine datasets relative to the name-masked reference.
\textbf{(c)} Correspondence-knowledge utility across a post hoc six-endpoint subset defined by whether the target itself is a matched variable, with the full 13-endpoint trajectory shown alongside it for comparison.
Intervals are 95\% CIs; metrics differ across panels and magnitudes are not directly comparable.
}
\label{fig:real_support_reference}
\end{figure}

\paragraph{Utility decreases as labeled target support increases.}
Labeled target examples give the reference another way to recover task-relevant information.
Accordingly, utility decreases as target support increases for the continuous clinical endpoints, in TabLLM, and the controlled benchmark (Figure~\ref{fig:real_support_reference}).

Among the six examination endpoints for which the target itself is a matched variable, correspondence utility falls from $+.070$ to $+.018$ as target support increases.
This attenuation is driven mainly by the reference catching up: SD-normalized RMSE improves from $.738$ to $.676$, while the reference improves from $.808$ to $.694$.
The broader 13-endpoint analysis shows the same downward trend, although utility at $n_{\mathrm{target}}=256$ remains unresolved ($+.014$, 95\% CI $[-.013,+.042]$).

The decline extends beyond continuous clinical endpoints.
TabLLM utility falls from $+.088$ to $+.007$, a decrease of $.081$ (95\% CI $[+.025,+.148]$), and utility is lower at $n_{\mathrm{adapt}}=512$ than at $0$ under all 24 evaluated references.
In the controlled benchmark, utility at the largest measurement mismatch falls from about $+.26$ to $+.03$ across the three prespecified target structures.

Together, these results show that utility is larger when the reference has fewer ways to recover the tested information. 
Making that information harder to recover increases utility, while labeled target examples allow the reference to catch up and reduce it.
Across both manipulations, the intended condition changes comparatively little; most of the variation comes from reference performance.
The measured utility also depends on the learner; additional capacity and learner-coverage analyses are reported in \supp{\S\ref{sec:app-relation-capacity}, \S\ref{sec:app-learner-coverage}}.

\subsection{Implications for Published Semantic Ablations}
\label{sec:broader_scope}

Given this dependence on the control, we ask whether published semantic ablations use controls that match their stated predictive claims.
Using a fixed inclusion rule, we audit 25 comparisons from nine studies that alter or remove an identifiable semantic component \supp{\S\ref{sec:published_audit}}.
This is a bounded audit of claim--control alignment, not an estimate of its prevalence in the literature.
For each comparison, we record the claim and assess whether the reported control isolates the claimed effect of the tested component.
For a predictive-utility claim, the control must remove the tested semantic component while preserving the remaining information, input interface, model, and learning procedure.
We classify the comparison as supported when all three requirements hold, unsupported when at least one clearly fails, and indeterminate when the published description is insufficient to establish them.

\begin{table}[!ht]
\centering
\vspace{-0.4em}

\caption{Claim--control alignment in the published-ablation audit.}
\label{tab:published_audit}
\vspace{0.1em}

\fontsize{8pt}{9pt}\selectfont
\setlength{\tabcolsep}{3pt}
\renewcommand{\arraystretch}{0.95}

\begin{tabular}{@{}lccc@{}}
\toprule
Claim type & Supported & Unsupported & Indeterminate \\
\midrule
Content sensitivity ($n=4$) & 3 & 1 & 0 \\
Predictive utility ($n=18$) & 1 & 11 & 6 \\
\bottomrule
\end{tabular}

\vspace{1pt}

{\scriptsize \emph{*Note.*} A comparison may make both claim types, so the rows are not mutually exclusive.
Counts include explicit claims only; for predictive utility, 18 comparisons make an explicit claim, five make no claim, and two have unclear scope.}

\vspace{-0.5em}
\end{table}

Among the 18 explicit predictive-utility claims, only one is paired with a control that satisfies all three requirements; 11 fail at least one requirement, and six cannot be determined from the published description.
Thus, most controls in this bounded sample do not by themselves isolate the predictive utility claimed for the tested semantic component.
Importantly, this does not show that the reported semantic knowledge is unhelpful.
It means only that the reported comparison does not isolate predictive utility under our criteria.

Because these judgments require manual coding, we also examined agreement between the two coders.
Preservation had the lowest agreement in the initial coding ($\kappa=.265$; \supp{\S\ref{sec:published_audit}}).
We therefore replaced the broad Preservation judgment with six fixed checks, after which the same two coders independently re-coded all 25 comparisons before reviewing disagreements.
Accordingly, the audit evaluates what each reported control can identify for the stated claim, rather than judging the overall correctness of the study or method.

\section{Conclusion}

Sensitivity to altered semantics does not by itself quantify the predictive contribution of semantic knowledge in cross-table transfer.
Across our analyses, predictive utility depends on the reference, the information available without the tested knowledge, labeled target support, and the model training setup.
Altered controls remain informative: they can show whether prediction depends on the supplied semantic content, but they should not be interpreted as direct estimates of predictive utility.
Our bounded audit shows that this distinction also matters in published practice, where many reported controls do not by themselves isolate the claimed predictive utility.
These results support claim-matched evaluation: controls should be chosen for the predictive claim they support, with predictive-utility claims requiring a suitable reference and reference sensitivity reported when several are available.

\section*{Data Use and Availability}

This study uses secondary data from NHANES, KNHANES, the Health and
Retirement Study (HRS), CAMELS-US, CAMELS-GB v2, and the nine public
tabular benchmark datasets used in the TabLLM reanalysis.

The controlled benchmark uses an eight-variable NHANES panel with source
cycles from 2001--2002 through 2013--2014 and target data from
2015--2016, 2017--2018, and the August 2021--August 2023 release; the
latter is labeled 2023 in our local preprocessing
\citep{cdcnhanesdata}. The real clinical-transfer experiments use
separately prepared NHANES and KNHANES inputs, so these cycle ranges
should not be interpreted as their participant-year ranges.

For KNHANES, the exact raw release and participant-year composition of
the frozen prepared input could not be fully reconstructed
\citep{kweon2014knhanes}.

For HRS, the associated preparation chain identifies the RAND HRS
Longitudinal File 2022 (V1), the 2016 Biomarker Data (Early v1.0), and
the 2016 Venous Blood Study (Final V3.0)
\citep{hrs2025randpublic,rand2025hrslong,hrs2020biomarker2016,hrs2025vbs2016};
exact byte-level lineage from these products to the frozen analysis
snapshot could not be fully reconstructed.

The hydrological experiments use CAMELS-US hydrometeorological time
series and catchment attributes
\citep{newman2014camelsdata,addor2017camelsattributes} and CAMELS-GB v2
\citep{coxon2025camelsgbv2}.

The accompanying repository does not redistribute participant-level
clinical source data, model weights, or row-level predictions.
Data requiring registration or supplemental authorization must be
obtained independently from the respective providers under their
applicable terms.

\section*{Reproducibility Statement}

The supplementary material provides the experimental settings, inference
procedures, analysis provenance, reference constructions, and robustness
analyses required to reproduce the reported comparisons.
The accompanying artifact contains code, frozen semantic records and
reference libraries, analysis manifests, environment information, and
mappings from reported tables and figures to their generating scripts
and outputs.
Data-access boundaries, product identifiers, provenance limitations, and
reproducible derived outputs are documented in the supplement and public
repository.

\section*{AI Use Statement}

Generative AI tools were used in this work in two roles.
First, language models were used in parts of the semantic-knowledge construction process to propose candidates from schema metadata and dataset documentation.
These steps did not use target labels or predictive performance, and the resulting semantic specifications were fixed before evaluation, as described in Section~\ref{sec:knowledge_construction}.
Second, generative AI tools were used during research development to provide feedback on experimental design and methodology, and during manuscript preparation to refine the phrasing and structure of the text. 
Their role in manuscript preparation was limited to improving clarity, coherence, and readability based on author-provided drafts. All scientific decisions, experimental implementation, analysis, and interpretation were conducted and verified by the authors.

\section*{Acknowledgments}

This work was supported by the Korean ARPA-H Project through the Korea
Health Industry Development Institute (KHIDI), funded by the Ministry of
Health \& Welfare, Republic of Korea (No. RS-2025-25455839).

We thank the Artificial Intelligence Industry Cluster Agency (AICA), Republic of Korea, for providing NVIDIA B200 GPU resources through its AI Data Center service.

The HRS (Health and Retirement Study) is sponsored by the National
Institute on Aging (grant numbers NIA U01AG009740 and NIA R01AG073289);
the study is conducted at the University of Michigan.
RAND HRS Data Products are produced by the RAND Center for the Study of
Aging with support from the National Institute on Aging and the Social
Security Administration.

This analysis uses Early Release data from the Health and Retirement
Study, (2016 Biomarker Data (Early v1.0) and the 2022 Early Release
component of the RAND HRS Longitudinal File 2022 (V1)), sponsored by the
National Institute on Aging (grant number NIA U01AG009740) and conducted
by the University of Michigan. These data have not been cleaned and may
contain errors that will be corrected in the Final Public Release
version of the dataset.

Contains data supplied by UK Centre for Ecology \& Hydrology, British
Geological Survey, Environment Agency, Natural Resources Wales and
Scottish Environmental Protection Agency.

\bibliography{references_seed}
\bibliographystyle{iclr2027_conference}

\ifdefined\ICLRIncludeAppendix
\ifdefined\ICLRIncludeAppendix
\def\ICLRFinishAppendix{ }
\else
\pdfminorversion=7
\documentclass{article}
\usepackage{iclr2027_conference,times}
\input{math_commands.tex}

\usepackage{url}
\usepackage{graphicx}
\usepackage{booktabs}
\usepackage{array}
\usepackage{longtable}
\usepackage{float}
\usepackage[section]{placeins}
\usepackage[hidelinks]{hyperref}
\newcolumntype{P}[1]{>{\raggedright\arraybackslash}p{#1}}
\setlength{\emergencystretch}{2em}

\title{Supplementary Material for\\
Beyond Correctness: Evaluating Semantic Knowledge in Cross-Table Transfer}
\author{Anonymous Authors}


\begin{document}

\maketitle
\def\ICLRFinishAppendix{%
\bibliography{references_seed}%
\bibliographystyle{iclr2027_conference}%
\end{document}%
}
\fi

\appendix
\raggedbottom
\ifdefined\ICLRArxivVersion
\renewcommand{\topfraction}{0.92}
\renewcommand{\bottomfraction}{0.85}
\renewcommand{\textfraction}{0.08}
\renewcommand{\floatpagefraction}{0.75}
\setcounter{topnumber}{4}
\setcounter{bottomnumber}{2}
\setcounter{totalnumber}{6}
\fi
\renewcommand{\thefigure}{A\arabic{figure}}
\renewcommand{\thetable}{A\arabic{table}}
\renewcommand{\theequation}{A\arabic{equation}}
\ifdefined\theHfigure
\renewcommand{\theHfigure}{appendix.\arabic{figure}}
\renewcommand{\theHtable}{appendix.\arabic{table}}
\renewcommand{\theHequation}{appendix.\arabic{equation}}
\fi
\setcounter{figure}{0}
\setcounter{table}{0}
\setcounter{equation}{0}

\section{Supplementary Experimental and Construction Details}
\label{sec:app-scope}

This appendix provides additional experimental details, extended results, and robustness analyses for the experiments in the main paper.
Unless stated otherwise, the reported results are specific to the evaluated datasets, learners, interventions, and references.
The public repository contains code, frozen semantic records and reference libraries, prompt and generation records for the LLM-based construction step, analysis manifests, environment information, and mappings from reported tables and figures to their generating scripts and outputs.

\subsection{Common experimental and inference settings}
\label{sec:app-experimental-details}

\paragraph{Comparisons and notation.}
Content sensitivity compares the intended and altered conditions, whereas predictive utility compares the intended condition with a suitable reference that removes the tested semantic knowledge while preserving the other predictive information.
Target-only models are reported separately when relevant to show performance without source transfer; they are not used as references for individual semantic components.

We use the following decomposition:

$$
Q_{\mathrm{intended}} - Q_{\mathrm{altered}}
=
\left(
Q_{\mathrm{intended}} - Q_{\mathrm{reference}}
\right)
+
\left(
Q_{\mathrm{reference}} - Q_{\mathrm{altered}}
\right).
$$

The first term on the right is predictive utility.
The second term shows how much of the intended--altered difference comes from the reference--altered difference; we do not interpret it as a separate effect of interest.

\paragraph{Scoring and reporting conventions.}
Intended--altered and intended--reference differences are reported so that positive values favor the intended condition.
The controlled benchmark uses
\[
Q=-\mathrm{MSE}/\mathrm{Var}(y^{\mathrm{qry}}).
\]

Absolute performance is reported as
\[
\mathrm{nMSE}
=
\mathrm{MSE}/\mathrm{Var}(y^{\mathrm{qry}})
=
-Q.
\]
Real-data experiments use their original AUROC, nRMSE, or log-RMSE metrics, so numerical effect sizes are compared only within the same metric.

Whenever nRMSE is reported, it is defined separately for each endpoint,
transfer direction, and split using the held-out target query set \(Q\):
\[
\operatorname{RMSE}_Q
=
\sqrt{
\frac{1}{|Q|}
\sum_{i\in Q}
(y_i-\hat y_i)^2
},
\qquad
s_Q
=
\sqrt{
\frac{1}{|Q|}
\sum_{i\in Q}
(y_i-\bar y_Q)^2
},
\]
and
\[
\operatorname{nRMSE}_Q
=
\frac{\operatorname{RMSE}_Q}{s_Q}.
\]
Here \(s_Q\) is the population standard deviation of the target labels in the
same held-out target query set (\(\mathrm{ddof}=0\)).
Query labels are used only for evaluation and normalization, not for model fitting.
Lower nRMSE is better; under the common higher-is-better convention,
\(Q=-\mathrm{nRMSE}\), so intended--reference utility is equivalently
\(\mathrm{nRMSE}_{\mathrm{reference}}
-\mathrm{nRMSE}_{\mathrm{intended}}\).

This definition is distinct from the controlled benchmark's primary
normalized error,
\(\mathrm{nMSE}=\mathrm{MSE}/\mathrm{Var}(y^{\mathrm{qry}})\).

\paragraph{Common experimental settings.}
Labeled target rows used for training are separate from the query rows, and query labels are used only for evaluation.
Within each comparison, conditions use the same data split, source sample, labeled-target sample, and random seed whenever applicable.

\begin{table}[t]
\caption{Common experimental settings across the real-data and controlled experiments.}
\label{tab:app-common-settings}
\centering
\footnotesize
\setlength{\tabcolsep}{2pt}

\begin{tabular}{@{}P{0.13\linewidth}P{0.18\linewidth}P{0.12\linewidth}P{0.225\linewidth}P{0.14\linewidth}P{0.115\linewidth}@{}}

\toprule
Setting & Target setup & Model & Model settings & Runs & Inference unit \\
\midrule

NHANES--\newline KNHANES &
Source rows plus \(n_{\mathrm{target}}=256\) labeled target rows; disjoint target query &
XGB &
300 trees, depth 6, learning rate .05; source and target total weights balanced &
10 seeds (50--59) per endpoint &
Endpoint \\
\midrule

NHANES/\newline KNHANES--HRS &
All eligible source rows (endpoint-specific); 1/5/10\% labeled target rows &
XGB &
300 trees, depth 5, learning rate .05; target weight 10 &
10 seeds (40--49) per labeled-target fraction &
Endpoint \\
\midrule

CAMELS-US--\newline CAMELS-GB v2 &
1/3/5 labeled target basins; fixed disjoint query basins &
XGB &
300 trees, depth 6, learning rate .05; log-target regression; target weight 10 &
20 seeds (20--39) per labeled-target size &
Basin \\
\midrule

Controlled &
2,048 source rows; \(n_{\mathrm{target}}\in\{32,512\}\); 2,000 query rows &
XGB, HistGB, TabPFN, MLP &
Primary conditions use unit weights; XGB/HistGB: 300 depth-6 iterations; TabPFN 7.0.0 defaults; MLP: one 32-unit ReLU layer &
20 realizations \(\times\) 3 fixed families &
Realization within family \\

\bottomrule
\end{tabular}
\end{table}

For the pooled NHANES/KNHANES--HRS relation experiments, all eligible
pooled source rows are used, with endpoint-specific source sizes of
113,229 (BUN), 122,588 (DBP), 90,060 (glucose), 87,466 (HbA1c),
58,737 (hematocrit), 126,500 (RBC), and 122,615 (SBP).
The corresponding HRS eligible totals before the 1/5/10\% labeled-target
split are 9,853, 26,932, 9,817, 6,486, 9,612, 9,615, and 26,932,
respectively.

Model-specific procedures for the retrospective TabLLM experiments and the
TransTab/CARTE learner-coverage extensions are reported in
Sections~\ref{sec:app-tabllm-reproduction} and
\ref{sec:app-learner-coverage}, respectively.
For conditions that use source data, source and labeled-target rows are concatenated and fit with the same model configuration.
Target-only models use only the labeled target rows and are reported separately when relevant.

\paragraph{Pairing and intervals.}
Conditions within the same comparison use the same data split and random seed, and query performance is never used to select hyperparameters.
Unless otherwise stated, bootstrap intervals use 10,000 resamples.
The paired reference-variation analysis in
Section~\ref{sec:app-tabllm-reference-noise} uses 5,000 resamples,
and the fixed-mapping lexical-reference analysis in Section C.3 uses
2,000 paired resamples.
For the survey experiments, we resample endpoints and their paired seeds; for hydrology, we resample basins; and for the controlled experiments, we resample the 20 realizations within each outcome family.
When an analysis uses a different resampling procedure or decision rule, we state it in that section.
Alternative interval estimates are reported only as robustness checks and do not replace the primary result.
The reported intervals apply to each contrast separately; we do not adjust them jointly across all analyses.
Accordingly, resolved and unresolved refer to the stated contrast rather than to a joint discovery claim.
When many references are compared at once, we use simultaneous intervals for the sign-change analysis.

\subsection{Analysis provenance}
\label{sec:app-registry}

We distinguish analyses by when their design was fixed relative to the results they evaluate.
\emph{Prespecified} analyses were fixed before the corresponding comparison;
\emph{frozen extensions} were added after earlier analyses but fixed before their new execution;
\emph{post-result} analyses were developed after relevant earlier results were known; and
\emph{retrospective} analyses operate on previously published or released results.
Replication and exploratory labels describe analysis scope rather than timing.
The complete analysis ledger, including construction dates and decision records, is retained in the public repository.

\begin{table}[h]
\caption{
Provenance information for analyses whose timing affects interpretation.
}
\label{tab:app-analysis-registry}
\centering
\footnotesize
\setlength{\tabcolsep}{3pt}

\begin{tabular}{P{0.30\linewidth}P{0.25\linewidth}P{0.37\linewidth}}
\toprule
Analysis & Timing / status & Interpretation \\
\midrule

Measurement endpoint extension
& Frozen extension
& The broader ten-endpoint result is descriptive. \\

Examination width-matched reference
& Post-result construction
& Enables a new predictive-utility comparison; it does not retroactively make the original ablation a suitable reference comparison. \\

Examination target-support ladder
& Direction calibrated earlier; six-endpoint panel selected after the 13-endpoint expansion
& Six-endpoint trajectory is treated as replication; the full 13-endpoint trajectory is exploratory. \\

Relation capacity analysis
& Frozen extension
& Evaluates learner-capacity dependence of relation utility. \\

Fixed-width noisy-proxy diagnostic
& Post-result diagnostic; protocol fixed before execution
& Tests reconstructibility while preserving input width and variable identity; the association within each noise level is post-result and descriptive. \\

Real-data reconstructibility probes
& Frozen extension
& The reconstruction-manipulation criterion was not met in the evaluated real-data settings. \\

TabLLM reproduction
& Retrospective compatibility reproduction
& Reanalyzes released predictions and code under a format-matched reference. \\

TabLLM reference-space audit
& Frozen extension
& The primary anonymous reference predates the audit; additional references evaluate sensitivity rather than reselecting the primary reference. \\

\bottomrule
\end{tabular}
\end{table}

\paragraph{Data products and provenance.}

The controlled benchmark and the real clinical-transfer experiments use
separate NHANES preparation paths; the controlled-benchmark cycle ranges
therefore do not describe the real-transfer input.

For KNHANES, linked metadata cover 1998, 2001, 2005, and 2007--2015,
but the exact raw release and participant-year composition of the frozen
prepared input could not be fully reconstructed.

For HRS, the associated preparation chain identifies the RAND HRS
Longitudinal File 2022 (V1), the 2016 Biomarker Data (Early v1.0), and
the 2016 Venous Blood Study (Final V3.0), but exact byte-level lineage
from these products to the frozen analysis snapshot could not be fully
reconstructed. The associated HRS preparation selects variable-specific
latest available values, so the HRS features and endpoints should not be
interpreted as measurements from a single visit.

The hydrological experiments use CAMELS-US hydrometeorological time
series and catchment attributes and CAMELS-GB v2 over the evaluated
period 1981-01-01 through 2008-09-30.

\subsection{Controlled benchmark construction}

\paragraph{Data and outcomes.}
The controlled benchmark uses an eight-feature NHANES panel while
preserving the observed feature values and missingness patterns.
Source participants are drawn from the 2001--2002 through 2013--2014
cycles and target participants from 2015--2016, 2017--2018, and the
August 2021--August 2023 release, which is labeled 2023 in local
preprocessing, with no participant overlap between the two groups.

Missing entries are filled only for synthetic-outcome construction using draws based on the observed source distribution; the learner continues to receive the original inputs with their original missingness.

For source-standardized mechanism terms $\widetilde h_t(z_i)$, outcomes are generated as
\begin{equation}
y_i = \eta_i + \epsilon_i,
\qquad
\eta_i = \sum_{t\in\mathcal T} w_t s_t \widetilde h_t(z_i),
\qquad
\epsilon_i \sim
\mathcal N\!\left(0,\operatorname{SD}_s(\eta)^2\right),
\end{equation}
where $w_t\sim U(.5,1)$ and $s_t\in\{-1,+1\}$ are fixed within each realization.
This noise construction gives a source signal-to-noise ratio of one.

We use three outcome families.
The additive family selects five single-feature terms.
The pairwise family selects three disjoint feature pairs over six features, and the sparse family selects two pairs from all 28 possible pairs.
Pairwise terms are either feature differences or log-differences,
with standardized inputs clipped to $[-5,5]$ before the logarithmic transformation.

Each realization fixes the outcome mechanism, schema representation, source and target rows, labeled-target sample, and noise before learner comparison.
We generate 20 realizations for each outcome family.

\paragraph{Schema differences and altered conditions.}
Four features are represented as ordinal variables and four as continuous variables.
Across the source and target schemas, ordinal features can differ in coding direction, code base, and missing-value code, while continuous features can differ in scale, offset, and missing-value code.
The target schema is kept fixed across all conditions.

For measurement knowledge, the altered condition changes the supplied coding information by permuting ordinal values, assigning continuous transformations to the wrong features, and changing the supplied missing-value code.
For correspondence knowledge, the altered condition reassigns features within the ordinal and continuous groups so that no feature remains matched to itself.
For relation knowledge, the altered condition replaces the intended relations with different relations while keeping the number and type of relations comparable.
Reversed and randomly replaced relations are evaluated separately.

\paragraph{Measurement mismatch level.}
For the measurement experiments, we also vary how many source features retain their dataset-specific schema.
Each realization uses a fixed ordering of the four ordinal and four continuous features.
At zero mismatch, all source features use the canonical representation.
As the mismatch increases, more source features use their dataset-specific representation, while the labeled target and query data remain unchanged.

\subsection{Evaluated feature and endpoint panels}
\label{sec:app-evaluated-panels}

\paragraph{Controlled feature panel.}
The controlled benchmark uses eight NHANES variables:
age, total cholesterol, HbA1c, creatinine, hemoglobin,
red blood cell count, white blood cell count, and waist circumference.
In each realization, four of the eight variables are randomly selected and
rendered as five-level ordinal variables using source quantiles, while the
remaining four are treated as continuous variables.
Thus, the ordinal subset is not tied to a fixed set of variable identities
and varies across realizations.

\paragraph{Survey-label measurement panel.}
The measurement experiments treat documented questionnaire codes as binary
prediction endpoints.
The focused panel contains diabetes history, hypertension history, and current
smoking status; the broader extension adds kidney disease, stroke, arthritis,
asthma, cancer, heart attack, and angina history.
The endpoint's own item is excluded from the predictor set.

\begin{table}[h]
\caption{
Survey-label endpoints used in the NHANES--KNHANES measurement experiments.
The first three rows form the focused panel; all ten rows form the extended panel.
}
\label{tab:app-measurement-endpoints}
\centering
\footnotesize
\setlength{\tabcolsep}{3pt}

\begin{tabular}{P{0.35\linewidth}P{0.22\linewidth}P{0.32\linewidth}}
\toprule
Endpoint & NHANES item & KNHANES item \\
\midrule
Diabetes history
& \texttt{DIQ010}
& \texttt{ALL\_\_de1\_dg} \\

Hypertension history
& \texttt{BPQ020}
& \texttt{ALL\_\_di1\_dg} \\

Current smoking status
& \texttt{SMQ040}
& \texttt{ALL\_\_bs3\_1} \\

Kidney disease history
& \texttt{KIQ022}
& \texttt{ALL\_\_dn1\_dg} \\

Stroke history
& \texttt{MCQ160F}
& \texttt{ALL\_\_di3\_dg} \\

Arthritis history
& \texttt{MCQ160A}
& \texttt{ALL\_\_dm1\_dg} \\

Asthma history
& \texttt{MCQ010}
& \texttt{ALL\_\_dj4\_dg} \\

Cancer history
& \texttt{MCQ220}
& \texttt{ALL\_\_dc1\_dg} \\

Heart attack history
& \texttt{MCQ160E}
& \texttt{ALL\_\_di5\_dg} \\

Angina history
& \texttt{MCQ160D}
& \texttt{ALL\_\_di6\_dg} \\

\bottomrule
\end{tabular}
\end{table}

\paragraph{Examination panel.}
The examination analysis contains 13 eligible endpoints:
glucose, waist circumference, triglycerides, systolic blood pressure,
diastolic blood pressure, total cholesterol, alanine aminotransferase (ALT),
blood urea nitrogen (BUN), creatinine, hemoglobin, hematocrit,
red blood cell count, and white blood cell count.
The original six-endpoint panel consists of glucose, waist circumference,
triglycerides, systolic and diastolic blood pressure, and total cholesterol.
The remaining seven endpoints were added in the broader endpoint extension.
Pulse rate was excluded by the frozen requirement of at least 1,024 eligible
target-domain rows.

The examination predictor registry contains 20 candidate slots.
The 11 base slots are age, creatinine, red blood cell count,
white blood cell count, hemoglobin, hematocrit, sex, education level,
diabetes history, hypertension history, and current smoking status.
The nine correspondence-member slots are height, weight, waist circumference,
systolic blood pressure, diastolic blood pressure, glucose, HbA1c,
total cholesterol, and triglycerides.
For each task, the target endpoint itself is removed from the predictor set.

\subsection{Additional details on suitable references}
\label{sec:app-reference-checks}

The main paper defines a suitable reference using three criteria: Removal, Preservation, and Setup.
Table~\ref{tab:app-reference-portability} shows how these criteria apply to the comparisons used in the measurement, correspondence, and relation experiments.
A comparison that fails one or more criteria may still be useful as an altered condition or descriptive ablation, but it is not used to estimate predictive utility.

\begin{table}[h]
\caption{
Application of the reference criteria across the measurement, correspondence, and relation experiments.
All listed comparisons satisfy the Setup criterion.
}
\label{tab:app-reference-portability}
\centering
\small
\setlength{\tabcolsep}{4pt}

\begin{tabular}{P{0.38\linewidth}ccP{0.34\linewidth}}
\toprule
Comparison & Removal & Preservation & Interpretation \\
\midrule

Measurement altered control
& $\times$
& \checkmark
& Altered condition \\
\midrule

Permuted correspondence
& $\times$
& \checkmark
& Content sensitivity only \\
\midrule

Survey ablation without added columns
& \checkmark
& $\times$
& Descriptive ablation \\
\midrule

Separate-domain reference
& \checkmark
& \checkmark
& Suitable reference \\
\midrule

Reversed / random relation control
& $\times$
& \checkmark
& Altered conditions \\
\midrule

Relation unconstrained reference
& \checkmark
& \checkmark
& Suitable reference \\

\bottomrule
\end{tabular}
\end{table}

The survey ablation without added columns satisfies Removal but not Preservation because it also removes other predictive information.
The permuted correspondence satisfies Preservation but not Removal because it changes the correspondence rather than removing it.
We therefore treat the former as a descriptive ablation and the latter as a content-sensitivity comparison.

\subsection{Semantic-knowledge construction and audits}
\label{sec:app-extraction-documentation}

\paragraph{Primary LLM-based candidate generation.}

For the NHANES--KNHANES documentation-based proposal run, we used GPT-5.6 Sol through Codex CLI 0.144.5 with reasoning effort set to \texttt{high}.
The system instruction and user template are stored separately for provenance, but during execution they were concatenated and sent as a single CLI input rather than as separate message roles.

The model receives schema metadata and selected documentation spans, but not person-level rows, target labels, predictions, or model-performance results.
The prompt also contains the fixed output JSON schema and domain-specific language (DSL) for executable relations.
Concept candidates may group multiple documented variables across the source and target schemas and are not restricted to one-to-one matches.
Relation candidates must use an operation from the fixed DSL and cite documentation for both source and target bindings.

After generation, fixed mechanical checks are applied to the output.
These checks cover JSON and schema validity, eligible columns, evidence references, fields or roles that could leak target information, operation and argument requirements, unit compatibility, duplicate candidates, and fixed limits on the number of candidates.
LLM confidence, downstream predictive performance, and human semantic judgments are not used to accept candidates.

The stored generation configuration lists greedy decoding, temperature 0, and seed 0.
These settings, however, do not appear as explicit options in the recorded Codex CLI command, so we do not claim deterministic decoding.
Instead, we preserve the system instruction, user template, rendered prompt, input and output schemas, validation rules, raw response, execution manifest, and cryptographic hashes for the executed run.
The experiments therefore use the stored candidate set from this recorded run; reproducibility does not require a new LLM call to generate the same candidates.

\paragraph{Seven-model measurement-code extraction audit.}

Separately from the candidate-generation procedure above, we evaluate measurement-code extraction from documentation across seven language models.
This audit uses its own prompt and validation procedure and is not a rerun of the GPT-5.6 Sol candidate-generation procedure.
Each model is run once with the fixed audit prompt, without external tools or retrieval, and the raw output is preserved.

Across seven evaluated models, five produced complete records in the required format, and four also produced valid-code sets that covered the target codes needed for evaluation.
The purpose of this analysis is to check whether measurement-code extraction can be carried out consistently from documentation alone, rather than to evaluate predictive utility.

\paragraph{Correspondence construction audit.}
\label{sec:app-correspondence-construction-audit}

Separately, we test correspondence construction in a larger candidate space containing 1,315 target variables.
Each of nine known source--target matches is evaluated against this full candidate set.

Using feature names alone recovers 2 of the 9 matches, while adding units recovers 4 of 9.
Using names, units, and codebook descriptions recovers all 9 matches.
With a single score threshold, 8 of the 9 true matches are retained in the standard setting.
When the true match is absent, the same threshold produces no false match in any of the nine no-match cases.

For comparison, the closed 9-by-9 matching task recovers all 9 matches.
This indicates that the smaller closed task provides an easier correspondence setting than matching against the full target schema.

\paragraph{Evidence for the relation set.}
\label{sec:app-relation-evidence}

The relations used in the real-data experiments were specified before predictive evaluation.
We subsequently reviewed prior literature to document the evidence for each relation and how closely that evidence matches the variables and prediction setting used here.
This review was used only to characterize the selected relations; it did not change the relation set.

\ifdefined\ICLRArxivVersion
\begin{table}[!htbp]
\else
\begin{table}[h]
\fi
\caption{External evidence for the relations used in the real-data experiments. The cited evidence supports the stated direction to different degrees and does not imply that each relation is a universal constraint.}
\label{tab:app-relation-evidence}
\centering
\footnotesize
\setlength{\tabcolsep}{3pt}

\begin{tabular}{P{0.20\linewidth}P{0.07\linewidth}P{0.27\linewidth}P{0.38\linewidth}}
\toprule
Relation & Dir. & Supporting evidence & Scope of the evidence \\
\midrule

Age $\rightarrow$ SBP
& $+$
& \citep{franklin1997hemodynamic,mitchell2010hemodynamic}
& Population trends do not imply the same association under every conditioning set. \\
\midrule

SBP $\leftrightarrow$ DBP
& $+$
& \citep{zhu2022pleiotropy,franklin1997hemodynamic}
& The population association is positive, although DBP changes differently with age. \\
\midrule

Fasting glucose $\leftrightarrow$ HbA1c
& $+$
& \citep{nathan2008translating}
& The cited relationship uses average rather than fasting glucose. \\
\midrule

Total cholesterol $\leftrightarrow$ triglycerides
& $+$
& \citep{friedewald1972estimation,wolska2026newmethods}
& Their joint use in LDL estimation is not direct evidence of a general population correlation. \\
\midrule

Weight $\rightarrow$ waist circumference
& $+$
& \citep{flegal2009comparisons,ross2020waist}
& Much of the cited evidence concerns BMI rather than weight alone. \\
\midrule

BUN $\leftrightarrow$ creatinine
& $+$
& \citep{hosten1990bun,morgan1977creatinine}
& Both reflect renal function, but their association is not uniformly strong across populations. \\
\midrule

Hemoglobin $\leftrightarrow$ hematocrit/RBC
& $+$
& \citep{doig2017methodical,guevara2023hemoglobin}
& The relationship depends on red-cell indices and population characteristics. \\
\midrule

Precipitation $\rightarrow$ discharge
& $+$
& \citep{zhang2001response,addor2017camels,coxon2020camelsgb,andreassian2025timeshift}
& Much of the cited evidence concerns annual or long-term behavior, whereas prediction uses daily data. \\
\midrule

PET $\rightarrow$ discharge
& $-$
& \citep{zhang2001response,addor2017camels,coxon2020camelsgb,andreassian2025timeshift}
& The long-term relationship is weak and not uniformly negative, and is applied here at a daily scale. \\

\bottomrule
\end{tabular}
\end{table}

\ifdefined\ICLRArxivVersion
\let\AppendixSavedFloatBarrier\FloatBarrier
\let\FloatBarrier\relax
\fi
\section{Additional Evidence on Content Sensitivity and Predictive Utility}
\ifdefined\ICLRArxivVersion
\let\FloatBarrier\AppendixSavedFloatBarrier
\fi

This section provides additional results comparing content sensitivity with predictive utility across the evaluated settings.

\subsection{TabLLM feature-name ablation}
\label{sec:app-tabllm-reproduction}

We revisit the published TabLLM feature-name ablation as an example in which the intended--altered difference can be compared with performance relative to a reference.

In the published zero-shot results, the mean intended--altered difference across the nine datasets is \(+.107\) AUROC.
The corresponding difference between the original List Template and the published values-only condition is \(+.076\).
The values-only condition, however, removes both feature names and the surrounding template structure, so we treat it as a published comparator rather than a suitable reference for feature-name utility.

The dataset-level results already show that the two differences need not agree.
For Credit-g, the intended--altered difference is \(+.09\), whereas the original names perform \(-.13\) below the values-only condition.
Thus, the large intended--altered difference in this case does not by itself show that feature names improve prediction.

To evaluate predictive utility while preserving the remaining input format, we reproduce the released TabLLM evaluation using a reference with fixed anonymous feature identifiers.
This reference keeps the List Template, field order, value formatting, and feature identity unchanged while replacing the feature names with identifiers such as \texttt{feature\_001}.
The identifiers keep the fields distinct without supplying their original feature meanings; differences among identifier strings are examined separately in Section~\ref{sec:app-tabllm-reference-audit}.

\paragraph{Compatibility reproduction.}
We treat this experiment as a released-code compatibility reproduction rather
than a bit-exact reproduction of the historical software stack.
The released 11B execution configuration and IA3 checkpoint resolve to
\texttt{bigscience/T0}, although one description in the original paper refers
to T0pp; we follow the executable released configuration and checkpoint.

The released repository contains only part of the corresponding T-Few
implementation, and the historical software stack no longer runs directly in
our environment.
We therefore use the model-modification code from the pinned
T-Few revision together with documented dependency and bfloat16 compatibility
fixes.
These changes preserve the released model initialization, serialization,
scoring procedure, and semantic comparison.

Under this compatibility reproduction, 21 of the 27 published-condition
dataset means are within .015 AUROC of the paper's printed values, with a
maximum absolute deviation of .0347.
We therefore do not claim a bit-exact numerical reproduction.
All semantic contrasts reported here are computed pairwise within this common
reproduction.

\paragraph{Adaptation protocol.}
We evaluate nine datasets---Bank, Blood, California Housing, Car, Credit-g,
Diabetes, Heart, Income, and Jungle---at
\(n_{\mathrm{adapt}}\in\{0,4,32,512\}\)
using the released split seeds
\(42, 1024, 0, 1, 32\).
The \(n_{\mathrm{adapt}}=0\) condition reuses the frozen zero-shot compatibility
reproduction without task-specific fitting.

For \(n_{\mathrm{adapt}}>0\), we initialize the released
\texttt{bigscience/T0} 11B model from the public IA3 checkpoint and update only
the 192 IA3 \texttt{lora\_b} tensors
(1,081,344 trainable parameters), while keeping the T0 backbone fixed.
We use the released T-Few objective consisting of the language-model,
multiple-choice, and unlikely losses.
Optimization uses Adafactor with learning rate \(0.003\), zero weight decay,
linear decay with 6\% warmup, gradient clipping at 1, batch size 4,
and 30 epochs.
This gives 30, 240, and 3,840 update steps for
\(n_{\mathrm{adapt}}=4,32,512\), respectively.
The maximum input length is 1,024 tokens.

For each seed, we construct the released-style 80/20 train--test split.
Adaptation examples are sampled class-balanced with replacement from the
remaining 80\%.
Paired semantic conditions use identical split membership, adaptation-row
membership, replacement multiplicities, row order, and loader seed.
Prediction uses normalized class-verbalizer sequence likelihoods;
binary tasks are evaluated by class-1 AUROC and Car by macro one-vs-rest AUROC.

The historical PyTorch-Lightning orchestration is replaced by a standalone
PyTorch compatibility port.
Sampling, tokenization, objective, optimizer, schedule, update count,
checkpoint initialization, and scoring follow the released T-Few setup and
were verified with a frozen zero-shot check before the
adaptation experiments.

For TabLLM, intervals use 10,000 paired multiplier resamples over the shared test examples, with dataset resampling for the macro estimate.
Zero-shot evaluations on the same test examples are treated as paired observations rather than independent runs.

\begin{table}[h]
\caption{
Released-code TabLLM reproduction using the anonymous-name reference.
Values are AUROC differences with 95\% paired intervals.
}
\label{tab:app-tabllm-reproduction}
\centering
\footnotesize
\setlength{\tabcolsep}{3pt}
\renewcommand{\arraystretch}{1.05}

\begin{tabular}{lrrr}
\toprule
Dataset & Content sensitivity & Predictive utility & Reference--altered \\
\midrule
Bank & -.032 [-.044,-.019] & +.089 [+.078,+.101] & -.121 [-.136,-.106] \\
Blood & +.038 [-.020,+.095] & +.114 [+.032,+.197] & -.076 [-.156,+.003] \\
California & +.084 [+.070,+.097] & +.053 [+.039,+.066] & +.031 [+.018,+.042] \\
Car & +.409 [+.370,+.448] & +.300 [+.268,+.333] & +.108 [+.068,+.149] \\
Credit-g & +.066 [+.005,+.129] & -.003 [-.047,+.039] & +.069 [+.015,+.124] \\
Diabetes & +.087 [+.053,+.122] & +.164 [+.102,+.224] & -.077 [-.138,-.016] \\
Heart & -.010 [-.067,+.048] & -.021 [-.051,+.009] & +.011 [-.040,+.063] \\
Income & +.139 [+.132,+.146] & +.074 [+.068,+.079] & +.066 [+.057,+.074] \\
Jungle & +.230 [+.221,+.240] & +.025 [+.017,+.032] & +.206 [+.197,+.214] \\
\midrule
Macro & +.112 [+.036,+.206] & +.088 [+.032,+.157] & +.024 [-.040,+.091] \\
\bottomrule
\end{tabular}
\end{table}

At the macro level, both content sensitivity and predictive utility are positive.
At the dataset level, however, the two measures can differ substantially.
Credit-g has positive content sensitivity (\(+.066\)) but approximately zero predictive utility (\(-.003\)), while Bank shows negative content sensitivity (\(-.032\)) despite positive predictive utility (\(+.089\)).
These cases illustrate that the intended--altered difference does not determine the predictive benefit of the intended semantic knowledge.

\subsection{Measurement knowledge}
\label{sec:app-measurement}

The focused three-endpoint analysis examines a setting in which coding polarity differs across schemas.
Without the intended decoding, reference performance falls below chance for diabetes and current smoking.
With the documented measurement information, both content sensitivity and predictive utility are positive.

Table~\ref{tab:app-measurement-full} also reports an extension to ten endpoints.
In this broader set, the reference AUROC is above chance on average, while predictive utility remains positive.
The intended--altered comparison, however, fails its frozen coverage requirement:
cancer history in KNHANES-to-NHANES transfer has only 4 of 10 finite seed pairs,
below the required eight. The frozen analysis retains the available finite pairs
within each endpoint--direction unit and weights endpoint clusters equally.
We therefore treat the ten-endpoint content-sensitivity result as descriptive
rather than as a coverage-qualified confirmation.

\begin{table}[h]
\caption{
Measurement-knowledge results with \(n_{\mathrm{target}}=256\).
Positive differences favor the intended condition.
The ten-endpoint content-sensitivity result is descriptive because its frozen
finite-pair coverage gate failed.}
\label{tab:app-measurement-full}
\centering
\footnotesize
\setlength{\tabcolsep}{3pt}

\begin{tabular}{
P{0.23\linewidth}
P{0.20\linewidth}
P{0.20\linewidth}
P{0.18\linewidth}
P{0.11\linewidth}}
\toprule
Panel
& Content sensitivity
& Predictive utility
& Reference--altered
& Reference AUROC \\
\midrule

Focused panel (3 endpoints)
& {+.292 [+.145, +.431]}
& {+.384 [+.202, +.553]}
& -.093 [-.228, +.012]
& .450 \\

Extended panel (10 endpoints)
& +.249 [+.182, +.309]
& {+.197 [+.110, +.294]}
& ---
& .556 [.499, .608] \\

\bottomrule
\end{tabular}
\end{table}

The focused panel shows positive content sensitivity and predictive utility under the polarity mismatch.
Predictive utility remains positive in the ten-endpoint extension, showing that the result is not limited to the two endpoints with below-chance reference performance, although the effect is smaller in the broader panel.

\subsection{Correspondence knowledge}
\label{sec:app-correspondence}

In the examination setting, the intended source--target correspondence outperforms a permuted correspondence with no correct matches, giving positive content sensitivity.
The original comparison, however, does not provide a suitable reference because removing the correspondence also changes the input representation.

We therefore construct a post-result width-matched reference with two columns for each correspondence member.
In the intended and altered conditions, source and target values assigned to the same correspondence share a column; the altered condition uses a permuted source--target assignment.
In the reference, source and target values instead occupy separate domain-specific columns, with rows from each domain empty in the other domain's column.

This construction preserves the base features, source transfer, labeled-target and query rows, and input width.
For each correspondence member, each row has exactly one designated value location before the original missingness pattern is applied.
Feature subsampling is disabled in all three conditions.
The reference therefore removes the correspondence between the source and target while preserving the other predictive information and input width.

Under the six checks used in the audit (Section~\ref{sec:published_audit}), this construction satisfies all six checks.
No source or target values or rows are removed; the same two columns are reserved for each correspondence member in every condition; and the model, preprocessing, and fitting procedure are unchanged.
The only change is whether corresponding source and target values share a column.

\begin{table}[h]
\caption{
Examination correspondence results with \(n_{\mathrm{target}}=256\).
The width-matched reference preserves the input width and underlying values while removing the source--target correspondence.
}
\label{tab:app-exam-no-bridge}
\centering
\footnotesize
\setlength{\tabcolsep}{3.5pt}
\renewcommand{\arraystretch}{1.05}

\begin{tabular}{P{0.18\linewidth}P{0.46\linewidth}P{0.22\linewidth}}
\toprule
Quantity & Comparison & Mean [95\% CI] \\
\midrule

Content sensitivity
& Intended -- altered correspondence
& {+.070 [+.029, +.121]} \\

Predictive utility
& Intended -- width-matched reference
& {+.047 [+.017, +.084]} \\

Reference--altered
& Width-matched reference -- altered correspondence
& +.022 [+.002, +.043] \\

\bottomrule
\end{tabular}
\end{table}
Because this reference was constructed after the original correspondence result was known, its predictive-utility comparison is treated as a post-result reference analysis rather than as part of the original ablation design.
Both content sensitivity and predictive utility are positive in the six-endpoint examination panel, but the intended--altered difference is larger.
The difference between them reflects the lower performance of the altered correspondence relative to the reference.

We also evaluate the same comparison across 13 eligible examination endpoints.
Content sensitivity remains positive at \(+.057\) (95\% CI \([+.025,+.093]\)), whereas predictive utility is smaller and unresolved at \(+.014\) (95\% CI \([-.013,+.042]\)).
Thus, the broader endpoint set preserves the distinction between sensitivity to the supplied correspondence and its predictive benefit.

\paragraph{Performance under incorrect correspondences.}
To test whether an altered correspondence can itself reduce performance, we progressively replace intended correspondences with plausible alternative matches.
AUROC loss increases as more incorrect matches are introduced and approaches the loss under the fully permuted correspondence
(Figure~\ref{fig:app-correspondence-decoy-damage}).

\begin{figure}[h]
\centering
\includegraphics[width=\linewidth]{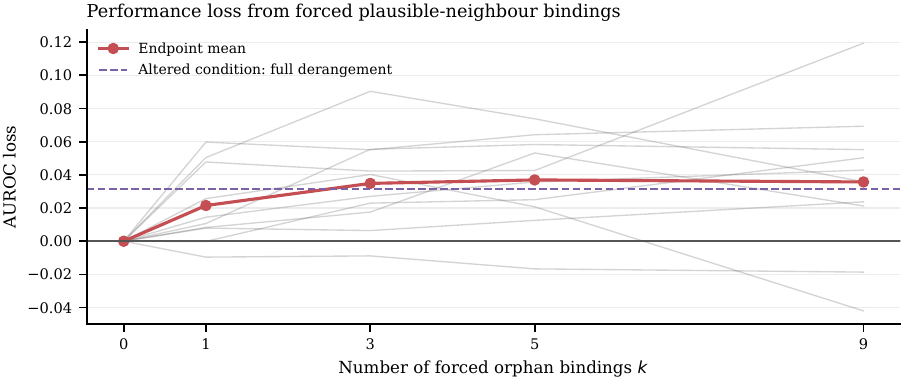}
\caption{
AUROC loss as intended correspondences are replaced by plausible alternative matches.
Thin lines show individual endpoints, the thick line shows their mean, and the dashed line shows the fully permuted correspondence.
}
\label{fig:app-correspondence-decoy-damage}
\end{figure}

This result provides direct evidence that part of an intended--altered correspondence gap can arise from degradation under the altered condition.

\subsection{Relation knowledge}
\label{sec:app-relation}

Relation knowledge shows a large difference between content sensitivity and predictive utility.
Across the real-data transfer settings, the intended relations outperform reversed relations by a substantial margin, while predictive utility relative to the unconstrained reference remains close to zero.

To test whether this difference depends on how the altered condition is constructed, we also replace the intended relations with randomly selected relations while keeping the intended and reference conditions fixed.
The intended--random differences are much smaller than the intended--reversed differences.

\begin{table}[H]
\caption{
Relation-knowledge results under reversed and random altered conditions.
Content sensitivity compares the intended relations with the indicated altered condition, while predictive utility compares the intended relations with the unconstrained reference.
Positive values favor the intended condition.
}
\label{tab:app-relation-altered-construction}
\centering
\footnotesize
\setlength{\tabcolsep}{3pt}

\begin{tabular}{P{0.30\linewidth}P{0.22\linewidth}P{0.22\linewidth}P{0.20\linewidth}}
\toprule
Setting
& Reversed sensitivity
& Random sensitivity
& Predictive utility \\
\midrule

NHANES--KNHANES
& +.183 [+.081, +.298]
& -.001 [-.006, +.003]
& -.002 \\

NHANES/KNHANES--HRS
& +.266 [+.175, +.367]
& +.041 [+.016, +.077]
& +.002 \\

CAMELS-120
& +.144 [+.127, +.163]
& +.002 [+.000, +.003]
& .000 [-.001, +.001] \\

\bottomrule
\end{tabular}
\end{table}

The contrast is strongest for the reversed condition.
For NHANES--KNHANES and CAMELS-120, replacing reversal with random relations reduces content sensitivity to approximately zero, while predictive utility is unchanged because the intended and reference conditions are identical across the two comparisons.
For HRS, the random relations also reduce performance relative to the reference, producing a smaller but still positive intended--random difference.

The controlled benchmark shows the same pattern when the intended relations are known by construction.
For the additive setting with \(n_{\mathrm{target}}=32\), content sensitivity is \(+.304\) against reversed relations but \(+.126\) against random relations, while predictive utility is only \(+.024\).
Thus, the magnitude of content sensitivity depends strongly on the altered relation used for comparison and need not reflect the predictive utility of the intended relation.

\section{Sensitivity to the Choice of Reference}

We next examine whether predictive utility changes across multiple suitable references.
We use the TabLLM feature-name experiment because feature-name information can be removed while keeping the remaining input format unchanged.

\subsection{Evaluated TabLLM reference constructions}
\label{sec:app-tabllm-reference-audit}

Starting from the anonymous-name reference used in
Section~\ref{sec:app-tabllm-reproduction},
we construct a frozen library of 24 anonymous-identifier references.
For a dataset with $p$ fields, every reference uses the same set of $p$
zero-padded identifiers
(\texttt{feature\_001}, \texttt{feature\_002}, $\ldots$)
and differs only in the one-to-one assignment of those identifiers to the fixed field lines.

The primary reference, denoted \texttt{ref\_00}, uses the original anonymous mapping.
The remaining 23 references use distinct non-identity permutations generated from a deterministic dataset-specific random stream.
Duplicate mappings are skipped, and the library size of 24 was fixed before scoring.

Across all candidates, the List Template, field order, values, placeholders, punctuation, and number of fields remain unchanged.
Each candidate also preserves a stable field-to-identifier mapping across rows.
Thus, the evaluated references differ only in which anonymous identifier is assigned to each feature.

For Blood, which contains four fields, the 24 references exhaust all $4!$ possible assignments.
For the other datasets, they form a fixed subset of the larger set of possible assignments.
Accordingly, the reported ranges characterize sensitivity over the evaluated library rather than bounds over all possible anonymous references.

Across the 24 evaluated references, mean predictive utility over the nine datasets ranges from
\(+.088\) to \(+.124\).
At the dataset level, the range is substantially wider and crosses zero for four of the nine datasets.

\begin{table}[h]
\caption{
Predictive utility across the 24 evaluated anonymous-name references.
Primary is the anonymous reference used in Section~\ref{sec:app-tabllm-reproduction}.
Minimum and maximum are descriptive point estimates across the evaluated references.
}
\label{tab:app-tabllm-reference-envelope}
\centering
\footnotesize
\setlength{\tabcolsep}{3pt}

\begin{tabular}{@{}lrrrr@{}}
\toprule
Dataset & Primary & Minimum & Maximum & Span \\
\midrule
Bank       & +.089 & +.022 & +.166 & .144 \\
Blood      & +.114 & +.093 & +.246 & .153 \\
California & +.053 & -.011 & +.105 & .116 \\
Car        & +.300 & +.270 & +.345 & .075 \\
Credit-g   & -.003 & -.040 & +.045 & .086 \\
Diabetes   & +.164 & +.088 & +.217 & .130 \\
Heart      & -.021 & -.021 & +.143 & .163 \\
Income     & +.074 & +.055 & +.106 & .051 \\
Jungle     & +.025 & -.020 & +.083 & .103 \\
\bottomrule
\end{tabular}
\end{table}

The point-estimate range crosses zero for California, Credit-g, Heart, and Jungle.
These ranges describe variation across the evaluated references; whether the apparent sign changes remain after accounting for evaluation uncertainty is examined in the next subsection.

\subsection{Reference variation and evaluation uncertainty}
\label{sec:app-tabllm-reference-noise}

The ranges in the previous subsection are based on point estimates.
We therefore test how much of the observed variation across references can be explained by evaluation uncertainty.

We use 5,000 paired bootstrap resamples over the shared test examples.
Within each resample, the same example weights are used for the intended condition and all 24 references, so uncertainty from the shared evaluation data is accounted for jointly.

Let $\hat{\mathbf u}$ denote the $R=24$ predictive-utility estimates for one dataset,
let $\widehat{\Sigma}$ denote their bootstrap covariance matrix, and define
\[
P = I-\mathbf{1}\mathbf{1}^{\top}/R.
\]
We measure the observed between-reference variance as
\[
V_{\mathrm{obs}}
=
\hat{\mathbf u}^{\top}P\hat{\mathbf u}/R,
\]
and estimate the contribution of evaluation noise to this spread as
\[
V_{\mathrm{noise}}
=
\operatorname{tr}(P\widehat{\Sigma})/R.
\]
The noise-corrected between-reference variance is therefore
$V_{\mathrm{obs}}-V_{\mathrm{noise}}$.
The projection removes common shifts shared across references, including uncertainty inherited from the common intended condition.

Across the nine datasets, 79.2--99.7\% of the observed between-reference variance remains after accounting for evaluation uncertainty.
Thus, most of the variation in utility across the evaluated references cannot be explained by test-set noise alone.

\begin{figure}[h]
\centering
\includegraphics[width=\linewidth]{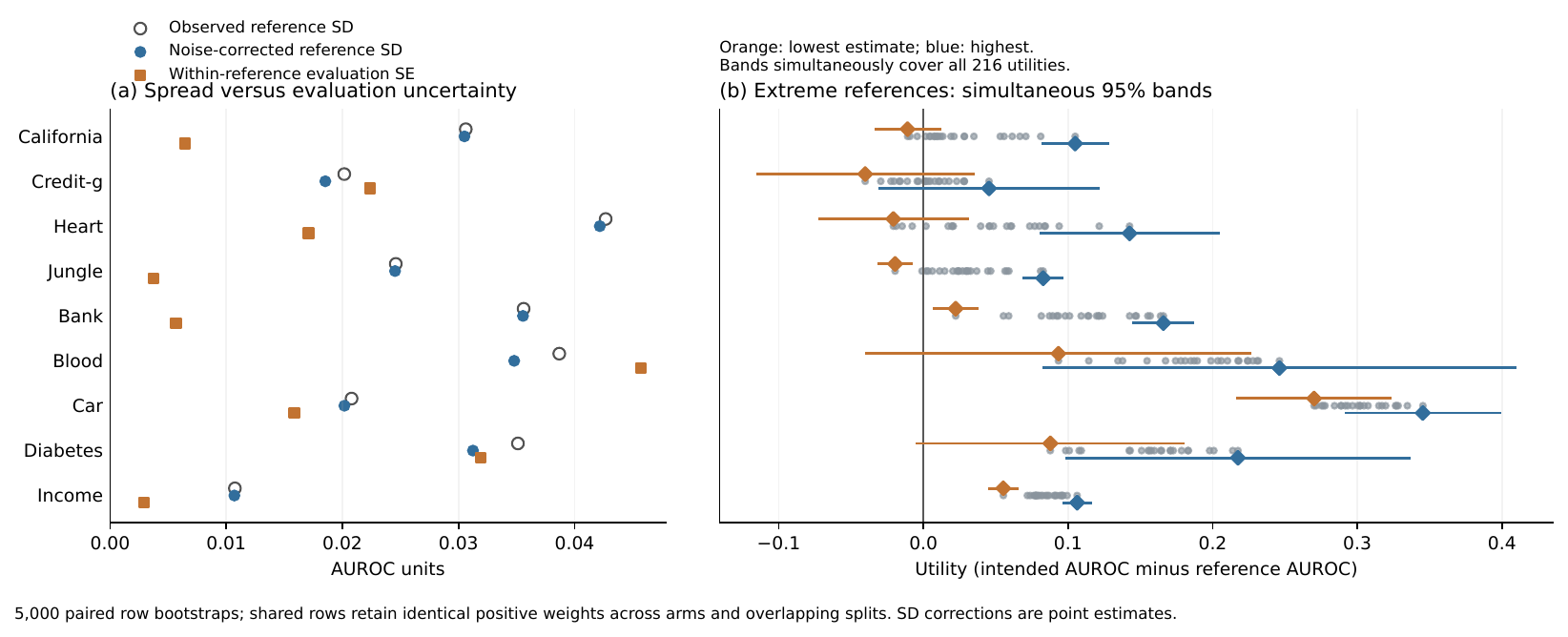}
\caption{
Reference variation and evaluation uncertainty across the 24 evaluated TabLLM references.
\textbf{(a)} Observed between-reference variation, its estimate after accounting for evaluation uncertainty, and the typical uncertainty of an individual reference estimate.
\textbf{(b)} Simultaneous 95\% intervals for the references with the lowest and highest estimated utility in each dataset; gray points show the remaining reference estimates.
}
\label{fig:app-tabllm-reference-noise}
\end{figure}

The point-estimate range crosses zero for California, Credit-g, Heart, and Jungle.
Only Jungle shows a resolved sign change: one simultaneous interval lies entirely below zero and another entirely above zero.
Its lowest estimated utility is \(-.0196\) (95\% CI \([-.0311,-.0081]\)), while its highest is \(+.0828\) (95\% CI \([+.0694,+.0962]\)).

For California, Credit-g, and Heart, at least one end of the range remains unresolved.
The 4/9 result therefore describes point-estimate sensitivity to reference choice, whereas a resolved change in sign is supported for 1/9 datasets.

\subsection{Additional reference checks}

\paragraph{Reproduction sensitivity.}
Because the released-code reproduction does not exactly match every published TabLLM result, we repeat the reference analysis on the four datasets for which the intended, permuted-name, and values-only conditions are all reproduced within .015 AUROC of the published means.
This restriction retains Car, Diabetes, Income, and Jungle and does not use the anonymous reference or predictive-utility results.

The main pattern remains in this subset.
Utility under the primary anonymous reference is \(+.141\), and the mean utility across the 24 evaluated references ranges from \(+.120\) to \(+.175\).
Dataset-level point-estimate ranges cross zero for one of the four datasets, compared with four of nine in the full panel.
Thus, positive aggregate utility and variation across references are not driven by the datasets with the largest reproduction differences.

\paragraph{Alternative anonymous labels.}
The primary reference uses artificial identifiers such as \texttt{feature\_001}.
To check whether the result depends on this particular wording, we repeat the zero-shot comparison using \texttt{Field A} and \texttt{Variable A} while keeping the feature mapping and the rest of the input format unchanged.

Mean predictive utility remains positive for all three versions: \(+.0887\) with \texttt{feature\_001}, \(+.0716\) with \texttt{Field A}, and \(+.0792\) with \texttt{Variable A}.
Relative to the primary reference, utility changes by \(-.0171\) for \texttt{Field A} (95\% CI \([-.0268,-.0077]\)) and by \(-.0095\) for \texttt{Variable A}.

These results show that the positive mean utility is not specific to the \texttt{feature\_001} labels, although its magnitude still depends on how the anonymous identifiers are written.

\section{Predictive Utility and Information Available to the Reference}

This section examines how predictive utility changes when the information supplied by semantic knowledge becomes easier or harder for the reference to recover.

\subsection{Measurement mismatch}
\label{sec:app-measurement-dose}

We vary the amount of measurement mismatch between the source and target schemas while keeping the target representation fixed.
Across all three outcome families and all evaluated models, predictive utility increases as more source features use their dataset-specific measurement representation
(Figure~\ref{fig:app-measurement-dose}).

With \(n_{\mathrm{target}}=32\), predictive utility at the largest mismatch is approximately \(+.26\) for XGB and HistGB and ranges from \(+.18\) to \(+.25\) for TabPFN.
With \(n_{\mathrm{target}}=512\), the corresponding effects are much smaller, ranging from \(+.02\) to \(+.04\).
For every model and outcome family, the change from \(n_{\mathrm{target}}=32\) to \(512\) is negative with a 95\% paired interval below zero.

\begin{figure}[h]
\centering
\includegraphics[width=\linewidth]{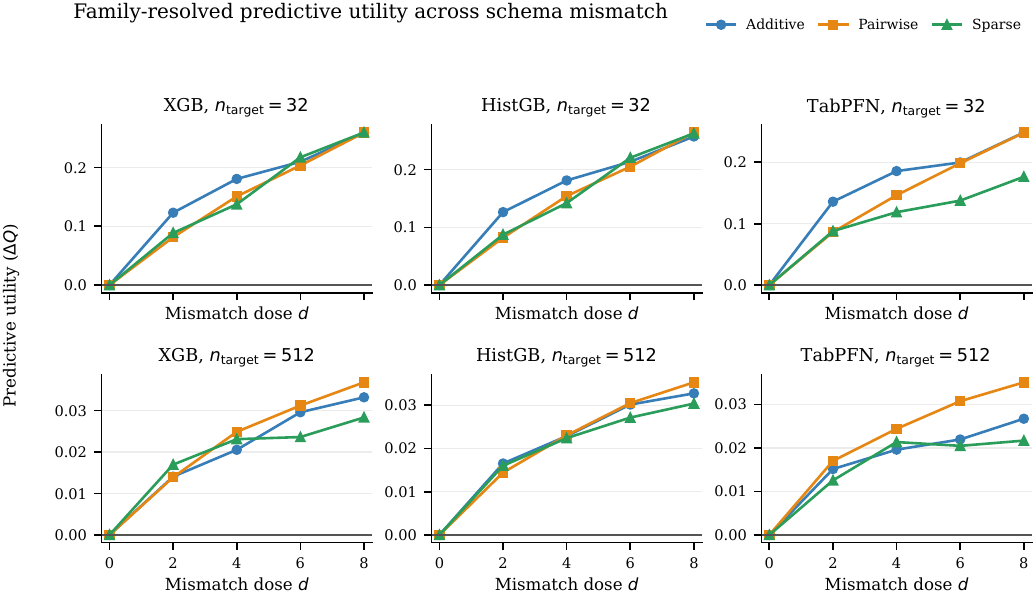}
\caption{
Predictive utility as measurement mismatch increases.
Curves show the mean over 20 realizations for the additive, pairwise, and sparse outcome families.
The upper row uses \(n_{\mathrm{target}}=32\), and the lower row uses \(n_{\mathrm{target}}=512\).
}
\label{fig:app-measurement-dose}
\end{figure}

\begin{table}[h]
\caption{
Summary of the measurement-mismatch experiment across the three outcome families.
Ranges report the minimum and maximum family-level values.
}
\label{tab:app-measurement-dose-stats}
\centering
\footnotesize
\setlength{\tabcolsep}{4pt}

\begin{tabular}{lccc}
\toprule
Model & \(n_{\mathrm{target}}\) & Spearman \(\rho\) & Full-mismatch utility \\
\midrule
XGB    & 32  & .78--.90 & +.259--+.260 \\
HistGB & 32  & .79--.89 & +.257--+.264 \\
TabPFN & 32  & .75--.85 & +.176--+.249 \\
XGB    & 512 & .55--.77 & +.028--+.037 \\
HistGB & 512 & .63--.76 & +.030--+.035 \\
TabPFN & 512 & .57--.80 & +.022--+.035 \\
\bottomrule
\end{tabular}
\end{table}

The same qualitative pattern appears across outcome families and model classes: greater measurement mismatch increases the value of the supplied measurement information, while additional labeled target data substantially reduces that value.

\subsection{Recoverability of relation values}
\label{sec:app-relation-value-interface}
We supply relation knowledge in two ways.
In the \emph{derived-value} setting, the model receives a value computed from the related variables.
In the \emph{directional-constraint} setting, the variables remain available and the relation constrains how the model may use them.

Here, recoverability means how well the relation value can be recovered from the remaining inputs.
The analyses below concern the derived-value setting.
They ask whether predictive utility increases when the supplied relation value becomes harder to recover from the remaining inputs.
They do not evaluate directional constraints.
A relation value and its negation can be equally recoverable from the same inputs, so recoverability does not determine the utility of a directional constraint.
We analyze directional constraints separately in Section~\ref{sec:app-relation-capacity} by comparing them with models that do not use these constraints.

When all constituent features are observed, adding the computed relation value provides almost no additional predictive utility.
We then progressively remove the constituent features used to compute that value while keeping the intended and reference conditions otherwise matched.
As the constituents are removed, predictive utility increases sharply.

For pairwise XGB with \(n_{\mathrm{target}}=32\), utility increases from \(+.002\) with all constituents present to \(+.242\) after all relevant constituents are removed.
The same pattern appears for the sparse family, both tree models, and both target sample sizes.
Across these settings, the intended condition changes comparatively little, while reference performance deteriorates as the relation becomes harder to recover from the remaining inputs.

\begin{figure}[h]
\centering
\includegraphics[width=0.9\linewidth]{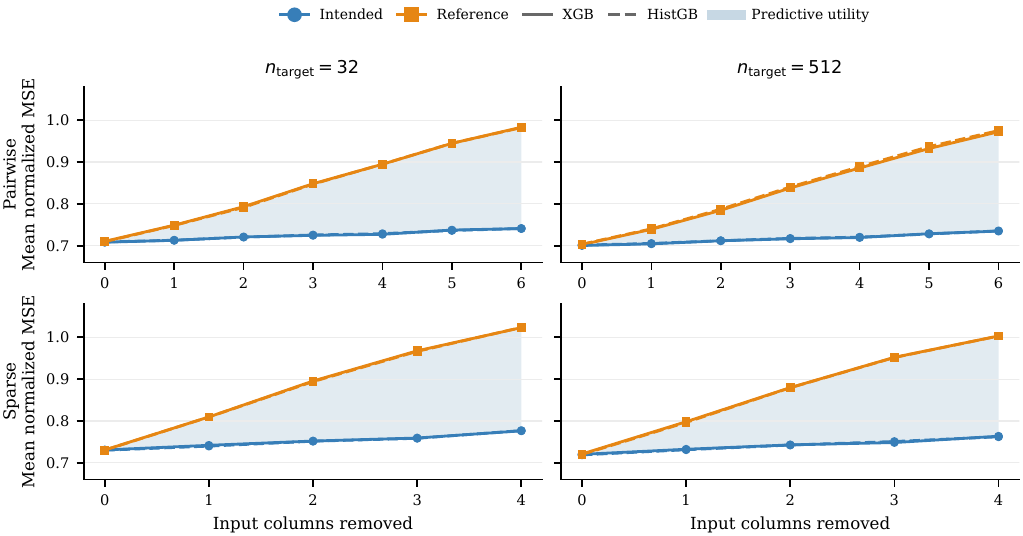}
\caption{
Performance as the constituent inputs of the supplied relation are progressively removed.
The intended condition changes comparatively little, whereas the reference deteriorates, increasing predictive utility.
Results are shown for the pairwise and sparse outcome families at two labeled-target sample sizes.
}
\label{fig:app-redundancy-ladder}
\end{figure}

This experiment changes both the available information and the number of input columns.
We therefore perform an additional experiment that keeps the input width fixed.
The constituent features remain present but are progressively corrupted with noise.

\paragraph{Recovery check with fixed input width.}
\label{sec:app-noisy-proxy}

For the pairwise and sparse outcome families, each constituent feature remains present under its original identity but receives Gaussian noise at
\[
\sigma\in\{0,.125,.25,.5,1,2,4\},
\]
with noise scaled by the source interquartile range of that feature.
At every noise level, the intended and reference conditions use the same eight input columns, noise values, rows, missingness pattern, data split, and learner.
Only the intended condition additionally receives the exact relation value computed from the clean constituent variables.

The experiment uses 20 realizations per outcome family, XGB and HistGradientBoosting, and
$n_{\mathrm{target}}\in\{32,512\}$.
The protocol was introduced after the earlier constituent-removal results were known but was fixed before this diagnostic was executed.
We therefore treat it as a post-result controlled diagnostic.
The recovery analysis was completed before predictive utility was evaluated.

As noise increases, the relation becomes harder to recover from the remaining inputs and predictive utility increases.
For representative pairwise XGB with \(n_{\mathrm{target}}=32\), reconstruction \(R^2\) falls from .987 to .032 while utility increases from \(+.002\) to \(+.266\).
For the sparse family, reconstruction \(R^2\) falls from .988 to .146 while utility increases from approximately zero to \(+.231\).

The same pattern appears across all eight combinations of outcome family, model, and target sample size.
At fixed nonzero noise levels, realizations with lower reconstruction $R^2$
generally have higher utility.
At each dose, we first compute Spearman correlation between $1-R^2$ and utility
across the 20 realizations separately in each family--learner--support setting,
then take the arithmetic mean across the eight settings.
These dose-specific means range from \(+.165\) to \(+.329\) across the six
nonzero doses.
This analysis within each noise level is post-result and descriptive.
Because it compares realizations at the same noise level, this analysis does not rely on the imposed ordering of noise levels and provides complementary evidence without changing input width or variable identity.

\paragraph{Limits of the real-data analysis.}
\label{sec:app-relation-real-data}

We also examined whether the same relationship could be isolated in real cross-table transfer.
The available real-data cases did not provide a clean test of how well the relation value could be recovered from the remaining inputs.
In CAMELS and NHANES--KNHANES, changing how well the value could be recovered also changed other input information, and HRS contained no eligible relation that could be supplied as a derived value.

We therefore treat this result as evidence from the controlled benchmark rather than as a mechanism established in the real-data transfer settings.

\subsection{Labeled target support}

Across three settings, predictive utility decreases as more labeled target data become available.
We examine this pattern for TabLLM feature-name information, controlled measurement knowledge, and correspondence knowledge in examination transfer.

\paragraph{TabLLM.}
Using the primary anonymous reference, feature-name utility decreases from \(+.088\) at
\(n_{\mathrm{adapt}}=0\) to \(+.007\) at \(n_{\mathrm{adapt}}=512\).

\begin{table}[H]
\caption{
TabLLM feature-name predictive utility as labeled adaptation data increase.
Intervals are 95\% dataset-bootstrap intervals.
}
\label{tab:app-tabllm-support-ladder}
\centering
\footnotesize
\setlength{\tabcolsep}{4pt}

\begin{tabular}{lrr}
\toprule
\(n_{\mathrm{adapt}}\) & Predictive utility & 95\% CI \\
\midrule
0   & +.088 & [+.034,+.153] \\
4   & +.073 & [+.012,+.144] \\
32  & +.035 & [+.017,+.054] \\
512 & +.007 & [+.003,+.011] \\
\midrule
Decrease, \(0\rightarrow512\) & +.081 & [+.025,+.148] \\
\bottomrule
\end{tabular}
\end{table}

The decrease is not specific to the primary reference.
Across all 24 anonymous references evaluated in
Section~\ref{sec:app-tabllm-reference-audit},
utility ranges from \(+.088\) to \(+.124\) at zero shot and from \(+.002\) to \(+.010\) at
\(n_{\mathrm{adapt}}=512\).
The corresponding decreases range from \(+.081\) to \(+.116\), and the bootstrap interval for the smallest decrease across the 24 references is
\([+.017,+.138]\).

The same pattern remains with alternative anonymous labels.
At \(n_{\mathrm{adapt}}=512\), mean utility is \(+.0090\) with
\texttt{Field A} and \(+.0083\) with \texttt{Variable A}, compared with
\(+.0071\) for the primary \texttt{feature\_001} reference.

The aggregate decrease masks substantial variation across datasets.
Seven of nine datasets have lower utility at \(n_{\mathrm{adapt}}=512\) than at zero shot,
but only two trajectories decrease monotonically across all evaluated support levels.

\ifdefined\ICLRArxivVersion
\begin{figure}[!htbp]
\else
\begin{figure}[H]
\fi
\centering
\includegraphics[width=0.82\linewidth]{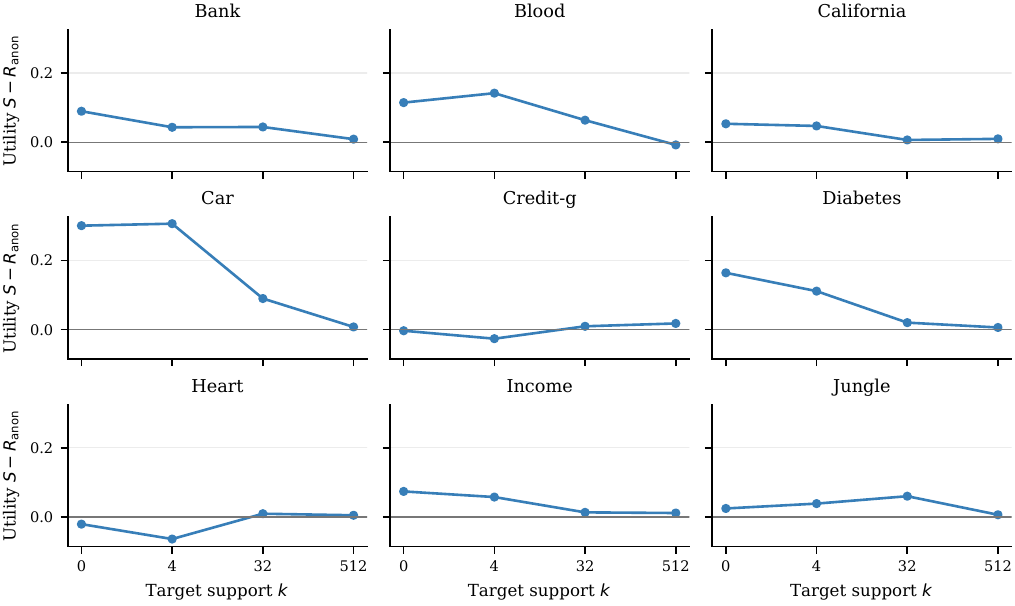}
\caption{
Dataset-level TabLLM feature-name utility across labeled adaptation levels.
Although mean utility decreases with additional labeled data, individual datasets show heterogeneous and often nonmonotonic trajectories.
}
\label{fig:app-tabllm-support-by-dataset}
\end{figure}

\paragraph{Controlled measurement knowledge.}
The controlled measurement experiment shows the same attenuation with labeled target support.
At the largest measurement mismatch, XGB predictive utility across the three outcome families decreases from \(+.259\)--\(+.260\) with
\(n_{\mathrm{target}}=32\) to \(+.028\)--\(+.037\) with
\(n_{\mathrm{target}}=512\).
The same qualitative pattern is observed for the other evaluated models
(Figure~\ref{fig:app-measurement-dose}).

\paragraph{Correspondence knowledge.}
In examination transfer, correspondence utility also decreases as the number of labeled target rows increases.
The six-endpoint subset was defined after the 13-endpoint expansion based on
whether the target was itself a matched variable, so we report the broader
13-endpoint analysis alongside it.

\begin{table}[h]
\caption{
Predictive utility of examination correspondence knowledge as labeled target data increase.
The six-endpoint panel contains targets that are themselves matched variables, while the 13-endpoint row reports the broader endpoint set.
}
\label{tab:app-exam-support-ladder}
\centering
\footnotesize
\setlength{\tabcolsep}{3pt}

\begin{tabular}{lrrrr}
\toprule
Endpoints
& \(n_{\mathrm{target}}=64\)
& \(n_{\mathrm{target}}=256\)
& \(n_{\mathrm{target}}=1024\)
& Decrease, \(64\rightarrow1024\) \\
\midrule

Six endpoints
& +.069 [+.037,+.102]
& +.047 [+.018,+.084]
& +.018 [+.008,+.028]
& +.052 [+.028,+.077] \\

All 13 endpoints
& +.039 [+.007,+.069]
& +.014 [-.013,+.042]
& +.008 [-.001,+.017]
& +.031 [+.003,+.056] \\

\bottomrule
\end{tabular}
\end{table}

In the six-endpoint panel, predictive utility decreases from \(+.069\) at
\(n_{\mathrm{target}}=64\) to \(+.018\) at \(n_{\mathrm{target}}=1024\).
The broader 13-endpoint analysis shows the same direction, although utility is unresolved at
\(n_{\mathrm{target}}=256\) and \(1024\).

Absolute performance improves in both conditions:
nRMSE decreases from .739 to .676 for the intended condition and from .808 to .694 for the reference.
The larger improvement of the reference accounts for the shrinking predictive utility.

The examination and controlled support manipulations differ in how labeled-target weight changes with sample size.
In the examination ladder, each labeled target row receives weight
$n_{\mathrm{source}}/n_{\mathrm{target}}$,
so the total target-domain training weight remains fixed while the number of distinct labeled target rows increases.
In the controlled benchmark, increasing $n_{\mathrm{target}}$ also increases the total target contribution because unit weights are used.
We therefore compare only the pattern of decreasing utility across the two settings rather than treating their support levels as quantitatively equivalent.

Across all three settings, additional labeled target information reduces the predictive utility of the supplied semantic knowledge.
The common pattern is that the reference improves as target information becomes more available, while the intended condition changes comparatively less.

\FloatBarrier

\section{Additional Robustness and Learner Coverage}

\subsection{Measurement robustness}

\paragraph{Polarity-invariant AUROC check.}
Because two reference AUROCs in the focused measurement panel are below chance,
we repeat the comparison using \(f(A)=\max(A,1-A)\).
Predictive utility remains positive in the three-endpoint panel,
\(+.276\) (95\% CI \([+.090,+.385]\)),
and in the ten-endpoint extension,
\(+.156\) (95\% CI \([+.089,+.225]\)).
Thus, the positive utility is not explained solely by AUROC polarity.

\paragraph{Rank-alignment check.}
The controlled measurement experiment combines differences in numerical representation with polarity reversals.
To separate these effects, we apply the same within-domain rank transformation to the intended and reference conditions before prediction.
This removes order-preserving differences in scale and coding while retaining polarity reversals.

With \(n_{\mathrm{target}}=32\), predictive utility remains positive in all six
outcome-family--model settings after rank alignment, with a mean of \(+.228\)
compared with \(+.260\) under the original representation.
With \(n_{\mathrm{target}}=512\), the remaining effect is much smaller,
ranging from \(+.008\) to \(+.019\) across the XGB outcome families.

After rank alignment, utility also increases with the number of reversed ordinal features.
At \(n_{\mathrm{target}}=32\), the mean increase is \(+.088\)
(95\% CI \([+.046,+.129]\)) per reversed feature.
At \(n_{\mathrm{target}}=512\), the corresponding slope decreases to \(+.004\)
(95\% CI \([+.001,+.008]\)).
When no ordinal feature is reversed, the rank-aligned intended and reference
inputs are identical, so predictive utility is exactly zero by construction.

\begin{table}[h]
\caption{
Summary of the controlled measurement robustness checks.
Rank-aligned utility removes order-preserving differences between the source
and target representations while retaining polarity reversals.
}
\label{tab:app-measurement-rank-alignment}
\centering
\footnotesize
\setlength{\tabcolsep}{4pt}

\begin{tabular}{lccc}
\toprule
Setting
& Original utility
& Rank-aligned utility
& Utility increase per reversal \\
\midrule

\(n_{\mathrm{target}}=32\), XGB
& +.259--+.260
& +.169--+.275
& +.088 [+.046,+.129] \\

\(n_{\mathrm{target}}=32\), HistGB
& +.257--+.264
& +.167--+.278
& +.088 [+.046,+.129] \\

\(n_{\mathrm{target}}=512\), XGB
& +.028--+.037
& +.008--+.019
& +.004 [+.001,+.008] \\

\bottomrule
\end{tabular}
\end{table}

Together, these checks show that the measurement result is not explained only by
below-chance AUROC or by order-preserving recoding.
After those effects are removed, the remaining utility is closely associated with
polarity mismatch and becomes much smaller when more labeled target data are available.

\subsection{Model and capacity dependence}

Relation utility varies with both model class and model capacity.
We first examine this dependence within the controlled benchmark and then extend the comparison to TransTab and CARTE.

\subsubsection{Tree capacity and MLP}
\label{sec:app-relation-capacity}

When the computed relation values are already available, adding the intended directional constraints provides a small positive gain across the tree-based settings.
Across the pairwise and sparse outcome families, this additional utility ranges from \(+.009\) to \(+.018\) over the two labeled-target sample sizes and the two tree models.

The magnitude of relation utility is sensitive to tree capacity.
Reducing XGBoost depth from 6 to 2 decreases utility by roughly an order of magnitude for the additive and pairwise families and makes the sparse-family mean slightly negative.

\begin{table}[H]
\caption{
Effect of XGBoost depth on relation predictive utility.
The baseline uses depth 6; the reduced-capacity model uses depth 2.
}
\label{tab:app-relation-capacity}
\centering
\footnotesize
\setlength{\tabcolsep}{4pt}

\begin{tabular}{lrrr}
\toprule
Family
& \(n_{\mathrm{target}}\)
& Depth 6
& Depth 2 \\
\midrule

Additive & 32  & +.0244 & +.0017 \\
Additive & 512 & +.0212 & +.0012 \\
Pairwise & 32  & +.0096 & +.0003 \\
Pairwise & 512 & +.0105 & +.0003 \\
Sparse   & 32  & +.0068 & -.0014 \\
Sparse   & 512 & +.0067 & -.0015 \\

\bottomrule
\end{tabular}
\end{table}

The MLP shows a different pattern.
Predictive utility is positive for the pairwise and sparse families, ranging from \(+.006\) to \(+.013\), but remains unresolved for the additive family.
The altered directional constraints also behave differently across families: they reduce performance in the additive setting but can outperform the reference in the pairwise and sparse settings.
Thus, both predictive utility and the intended--altered difference depend on the model through which the relation information is used.

\subsubsection{TransTab and CARTE}
\label{sec:app-learner-coverage}

We further evaluate the same distinction with TransTab and CARTE.
The results are not uniformly positive across models, semantic components, or labeled-target sample sizes.

\paragraph{TransTab setup.}
We use the official TransTab implementation at commit
\texttt{fdb34cf38abda73ee6a741b802fe226cc89ba7b5}
with a \texttt{TransTabRegressor} initialized from scratch.
The model uses hidden dimension 128, two Transformer layers,
eight attention heads, a 256-dimensional feed-forward layer,
ReLU activations, and zero dropout.
We train for 50 fixed epochs with Adam, learning rate \(10^{-4}\),
zero weight decay, and batch size 64, without validation-based model selection.

Source and labeled-target tables are provided jointly through TransTab's
variable-column supervised interface.
Training labels are standardized using the mean and population standard
deviation of the source training labels and transformed back to the original
scale for evaluation.
Numeric missing values are filled using training medians from each domain with
explicit missingness indicators; target-query preprocessing uses statistics
computed from the labeled target data.
The realization index is used as the random seed and is held fixed across
paired semantic conditions.

Two compatibility corrections were required for regression execution:
we corrected the upstream \texttt{TrainDataset.\_\_getitem\_\_} slice from
\texttt{index-1:index} to \texttt{index:index+1}, and read the raw regression
logit rather than the classifier-sigmoid output returned by
\texttt{predict()}.
Neither change modifies the architecture, training objective, or optimizer.

\paragraph{CARTE setup.}
We use \texttt{carte-ai==0.0.26} with a single
\texttt{CARTEMultitableRegressor}.
The distributed pretrained weights and fastText embeddings are loaded,
with the pretrained representation weights kept frozen.
Models are fit with learning rate \(0.001\), batch size 64,
a maximum of 200 epochs, validation fraction \(0.2\),
early-stopping patience 20, and target fraction \(0.125\).
The realization index is used as the paired random state.

As for TransTab, numeric missing values are processed using training medians
from each domain and explicit missingness indicators, and target-query
preprocessing uses statistics computed from the labeled target data.
Because CARTE does not support the directional constraints used by the tree
learners, relation knowledge is supplied through derived value-plus-text columns.
The CARTE result therefore tests whether the pattern extends to CARTE under
this derived-value representation; it is not a direct replication of the
tree-based directional constraint.

The exact upstream CARTE Git revision was not recorded in the frozen artifact;
we therefore anchor reproducibility to
\texttt{carte-ai==0.0.26} and the hashes of the distributed pretrained and
fastText assets retained in the artifact.

\begin{table}[h]
\caption{
Additional model coverage.
For CARTE, counts report the number of outcome families with a positive predictive-utility interval out of three evaluated families.
}
\label{tab:app-model-coverage-compact}
\centering
\footnotesize
\setlength{\tabcolsep}{4pt}

\begin{tabular}{P{0.20\linewidth}P{0.24\linewidth}P{0.48\linewidth}}
\toprule
Model / support
& Component
& Result \\
\midrule

TransTab / \(n_{\mathrm{target}}=32\)
& Shared cross-domain identity
& Utility \(+.114\)--\(+.129\); 74--80\% of the intended--altered gap is associated with altered-condition degradation \\

TransTab / \(n_{\mathrm{target}}=512\)
& Shared cross-domain identity
& Utility \(+.007\)--\(+.009\); 97--100\% of the intended--altered gap is associated with altered-condition degradation \\

CARTE / \(n_{\mathrm{target}}=32\)
& Measurement / correspondence / relation
& Positive utility in 0/3, 0/3, and 0/3 outcome families, respectively \\

CARTE / \(n_{\mathrm{target}}=512\)
& Measurement / correspondence / relation
& Positive utility in 2/3, 2/3, and 0/3 outcome families, respectively \\

\bottomrule
\end{tabular}
\end{table}

For TransTab, shared anonymous identifiers outperform domain-specific anonymous identifiers, whereas original feature names do not show a resolved improvement over the shared identifiers.
Because the regression head is newly initialized, this comparison supports the contribution of shared cross-domain identity rather than pretrained lexical semantics.

For CARTE, positive utility appears for measurement and correspondence knowledge in some \(n_{\mathrm{target}}=512\) outcome families, while relation utility remains unresolved.
Together with the tree-capacity and MLP results, these analyses show that the measured predictive utility of semantic knowledge depends on the model and how that knowledge is incorporated.

\subsection{Fine-grained and unit-level analyses}
\label{sec:app-controlled-boundaries}

\paragraph{How correspondence is supplied.}
Correspondence sensitivity depends on how the correspondence is supplied.
In the examination analysis, one representation explicitly pairs source and target columns.
The other aggregates columns by variable without separately encoding the source--target pairing.
The intended--altered difference is \(+.065\) with explicit source--target pairing but approximately zero under the aggregated representation.
In the survey analysis, the intended condition pairs the added source and target columns, while the altered condition permutes these pairs.
The intended--altered difference is positive with both XGB (\(+.045\)) and HistGB (\(+.051\)).
The survey ablation without added columns is reported only as an ablation because it removes additional input information and does not satisfy Preservation.

\paragraph{Individual CAMELS relations.}
The CAMELS content-sensitivity result is driven primarily by the precipitation direction.
Flipping precipitation increases log-RMSE by \(+.191\) (95\% CI \([+.143,+.248]\)), whereas flipping PET has approximately no effect,
\(-.001\) (95\% CI \([-.003,+.001]\)).
When both relations are considered together, the corresponding contributions are \(+.188\) for precipitation and \(-.003\) for PET.
Thus, the aggregate CAMELS result should not be interpreted as evidence that both declared relations contribute equally.

\paragraph{Resolution of unresolved effects.}
Zero-containing intervals differ substantially in precision.
For the clinical relation comparisons, the 80\% minimum detectable effect is
.0095--.0122 nRMSE, compared with .0015--.0029 log-RMSE for CAMELS.
The CAMELS-120 utility estimate is approximately zero with a narrow 95\% interval of
\([-.0011,+.0012]\) log-RMSE.
The three-endpoint survey correspondence comparisons have substantially lower resolution, with minimum detectable effects of .1443--.1586 AUROC.
Because these experiments use different metrics, the magnitudes are not compared directly, and the narrow CAMELS interval is not interpreted as formal equivalence to zero.

\paragraph{Unit-level heterogeneity.}
Aggregate results also hide variation across individual endpoints and basins.
Relation content sensitivity is positive across all evaluated HRS endpoints and the original CAMELS basins, whereas predictive utility varies around zero.
Survey correspondence results are more heterogeneous across endpoints.

\begin{table}[h]
\caption{
Unit-level variation in the main real-data comparisons.
Ranges summarize descriptive endpoint- or basin-level effects.
}
\label{tab:app-unit-attribution-summary}
\centering
\footnotesize
\setlength{\tabcolsep}{3pt}

\begin{tabular}{
P{0.27\linewidth}
P{0.14\linewidth}
P{0.26\linewidth}
P{0.26\linewidth}}
\toprule
Setting & Units & Content sensitivity & Utility / ablation \\
\midrule

Survey correspondence / XGB
& 3 endpoints
& all positive; +.022--+.072
& -.010--+.076 \\

Survey correspondence / HistGB
& 3 endpoints
& all positive; +.025--+.089
& -.024--+.073 \\

NHANES--KNHANES relation
& 6 endpoints
& all positive; +.010--+.445
& -.013--+.007 \\

NHANES/KNHANES--HRS relation
& 7 endpoints
& all positive; +.068--+.531
& -.014--+.019 \\

CAMELS relation
& 12 basins
& all positive; +.032--+.463
& -.004--+.008 \\

\bottomrule
\end{tabular}
\end{table}

\subsection{Exploratory cross-component interaction}
\label{sec:app-composition}

As an exploratory analysis, we examine whether measurement and correspondence
knowledge interact rather than contributing independently.
Let \(Q_{mc}\) denote performance for \(m,c\in\{0,1\}\), where 1 denotes the intended condition and 0 the altered condition.
We define

\[
I_{MC}
=
Q_{11}-Q_{10}-Q_{01}+Q_{00}.
\]
This interaction concerns the intended--altered conditions and is not an interaction of predictive utility relative to a reference.

In the controlled benchmark, the interaction is positive at low labeled-target
support and decreases as an increasing fraction of the source labels is changed while the input data remain fixed.
The initial real-data probes make the same change to the source labels but do not show the same decrease
(Figure~\ref{fig:app-interaction-controlled-real}).

\begin{figure}[H]
\centering
\includegraphics[width=\linewidth]
{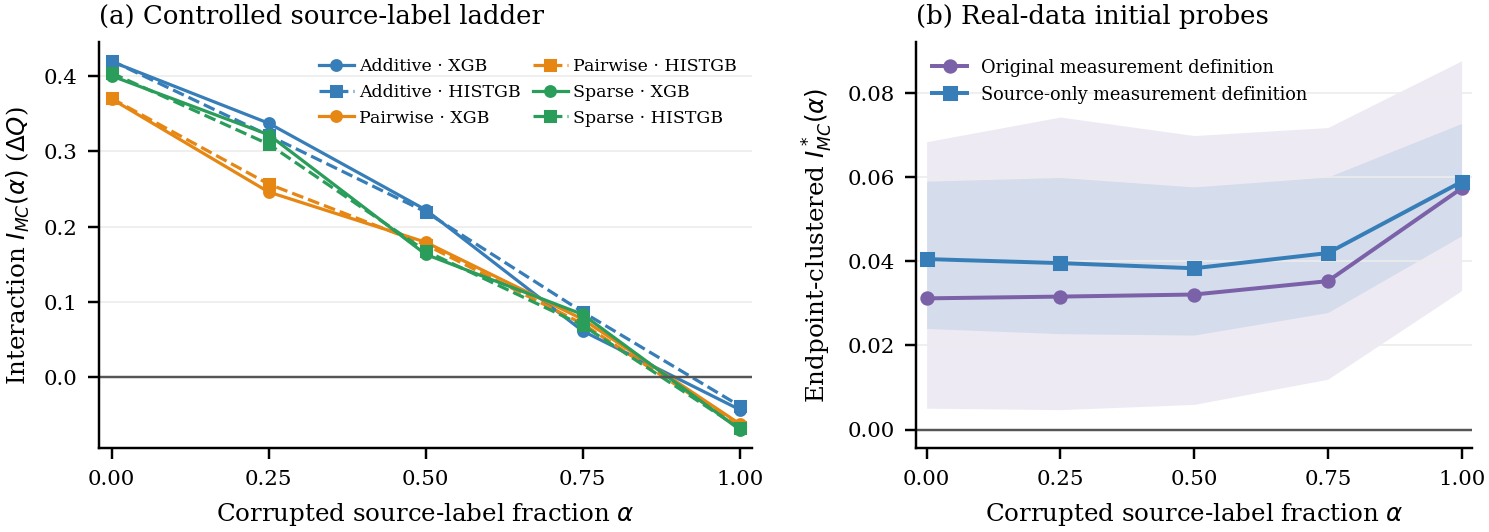}
\caption{
Exploratory measurement--correspondence interaction analysis.
\textbf{(a)} In the controlled benchmark, interaction decreases as an increasing fraction of the source labels is changed.
\textbf{(b)} The initial real-data probes do not show the same decrease.
}
\label{fig:app-interaction-controlled-real}
\end{figure}

A later real-data check considers three changes.
Target-label corruption changes the outcome labels of the target rows used for training.
Source rendering changes only how the source values are represented, and source-column permutation changes the assignment of source variables to columns while leaving the target columns unchanged.
The interaction decreases under target-label corruption, but not under the other two changes.
Because the result depends on which component is changed, we treat this analysis as
exploratory rather than evidence for a general cross-component mechanism.

\section{Published Semantic-Ablation Audit}
\label{sec:published_audit}

\subsection{Selection and coding procedure}

\paragraph{Study selection.}
We conduct a bounded retrospective audit of published semantic-ablation experiments rather than a systematic literature review.
The candidate pool consists of cross-table and semantic tabular-learning studies examined in our related-work review, together with closely related frameworks considered during revision.
A study is included if it reports at least one predictive comparison in which an identifiable semantic component supplied to the model is explicitly altered or removed.
Comparisons that vary pretraining, architecture, optimization, or representation mechanisms while leaving the evaluated semantic component unchanged do not satisfy this criterion.
A comparison can still enter the audit if it removes or alters the tested semantic component while also changing another part of the method; those additional changes are then evaluated under the criteria below.

For each included study, we retain all reported comparisons that satisfy this rule rather than selecting comparisons based on their effect direction or magnitude.
We additionally screened XTab and CM2 during revision.
Among the excluded studies, TransTab uses semantic column and categorical descriptions but does not report an ablation that alters or removes them; XTab does not use feature-level semantic conmponent as a model input; and CM2 uses column-name semantics but likewise reports no ablation that alters or removes them.
Table~\ref{tab:app-audit-screening} summarizes the screening decisions.
We screened 12 candidate studies and retained nine, yielding 25 qualifying comparisons.

\begin{table}[h]
\caption{Study screening for the published semantic-ablation audit.}
\label{tab:app-audit-screening}
\centering
\footnotesize
\setlength{\tabcolsep}{3pt}

\begin{tabular}{@{}P{0.20\linewidth}P{0.12\linewidth}P{0.62\linewidth}@{}}
\toprule
Study & Included & Basis for decision \\
\midrule
LIFT       & Yes & Explicit feature-name semantic manipulations \\
TransTab   & No & Uses column and categorical descriptions, but no reported ablation alters or removes this semantic knowledge \\
TabLLM     & Yes & Explicit name/value semantic manipulations \\
PLATO      & Yes & Explicit knowledge-graph ablations \\
CARTE      & Yes & Explicit semantic-representation ablations \\
FeatLLM    & Yes & Explicit feature-description ablation \\
TabuLa-8B  & Yes & Explicit header-semantic comparison \\
ConTextTab & Yes & Explicit semantic-representation/name ablations \\
TabSTAR    & Yes & Explicit verbalization ablations \\
TARTE      & Yes & Explicit semantic-encoding comparison \\
XTab       & No & Does not use feature-level semantic component as a model input \\
CM2        & No & Uses column-name semantics, but no reported ablation alters or removes them \\
\bottomrule
\end{tabular}
\end{table}
Our post-result TransTab experiments are separate learner-coverage analyses and are not part of the published evidence used to determine this study's audit eligibility.

\paragraph{Reported claims.}
For each comparison, two human coders (one author and one independent researcher) independently examined the reported result, its caption, and the surrounding interpretation.
They recorded whether the study explicitly used the comparison to support a content-sensitivity claim, a predictive-utility claim, both, or neither.
Ambiguous cases were coded as unclear.

A content-sensitivity claim states that prediction depends on the particular semantic content supplied.
A general statement that semantic knowledge is used or matters is not sufficient.
A predictive-utility claim states that the tested semantic knowledge improves predictive performance.

\paragraph{Control support.}
The coders then independently assessed what each reported control can establish.

For content sensitivity, the control must change the tested semantic content while preserving how that content is used and keeping the rest of the comparison unchanged.
For predictive utility, we apply the three criteria defined in the main paper: Removal, Preservation, and Setup.
Because Preservation had lower agreement in the initial coding, both coders also applied six fixed checks to all 25 comparisons: other predictive values (P1), feature set and identity (P2), input width (P3), input format (P4), model (P5), and learning and inference procedure (P6).
Each check was coded Yes, No, or Unclear from the paper and supplement, and the source passage was recorded.
Preservation requires P1--P4 to be Yes, and Setup requires P5--P6 to be Yes.
For either criterion, a No on any required check causes that criterion to fail; otherwise, any required Unclear makes it Unclear.
The six checks were fixed before the coders applied them.

A predictive-utility control is coded as \emph{Supported} when all three criteria are satisfied, \emph{Unsupported} when at least one criterion clearly fails, and \emph{Indeterminate} when the published description does not provide enough information to establish all three criteria.

We then compare the reported claim with what its control supports.
An explicit claim is supported when the corresponding control meets its requirements, unsupported when a required condition clearly fails, and indeterminate when the available description is insufficient.
Comparisons that do not make the corresponding claim are not counted as unsupported claims, and unclear claim statements are kept separate.

\subsection{Coding agreement}

Two human coders independently coded all 25 comparisons before any discussion. Table~\ref{tab:app-published-audit-agreement} reports agreement for this initial coding.

\begin{table}[H]
\caption{
Agreement between the two human coders before resolving disagreements.
}
\label{tab:app-published-audit-agreement}
\centering
\small
\setlength{\tabcolsep}{5pt}

\begin{tabular}{lcc}
\toprule
Coding item & Agreement & Cohen's \(\kappa\) \\
\midrule
Content-sensitivity claim   & 25/25 (100\%) & 1.000 \\
Predictive-utility claim    & 24/25 (96\%)  & .907 \\
Content-sensitivity support & 24/25 (96\%)  & .815 \\
Removal                     & 25/25 (100\%) & 1.000 \\
Preservation                & 15/25 (60\%)  & .265 \\
Setup                       & 22/25 (88\%)  & .784 \\
Predictive-utility support  & 19/25 (76\%)  & .649 \\
\bottomrule
\end{tabular}
\end{table}

Preservation had the lowest agreement in the initial coding, with agreement on 15 of 25 comparisons ($\kappa=.265$).
The same two coders then applied the fixed P1--P6 checks independently to all 25 comparisons and recorded the source passage for each answer. They agreed on 105 of the 150 individual checks (70.0\%). Across the 18 explicit predictive-utility claims, the support labels derived from these independent checks agreed in 15 cases (83.3\%, $\kappa=.693$).
After both coding sheets were complete, disagreements and cases flagged by a consistency check were reviewed against the source passages.
Disagreements were resolved by consensus.
The final consensus codes are not treated as an inter-rater agreement measure. The full P1--P6 coding and source passages are included in the artifact.

A standalone contemporaneous record of the final C01-B content-sensitivity
adjudication was not retained. During artifact finalization, both coders
re-reviewed the original independent sheets and cited source evidence and
confirmed the reported final label; this finalization record is included in
the artifact.

AI tools were used only to organize cases and locate the relevant source passages. The coders made the judgments, and the authors reviewed the final coding.

\subsection{Full audit results}

Table~\ref{tab:app-published-audit} reports the comparison-level coding for all 25 audited ablations.
For each claim type, we show whether the original study made the claim and whether the reported control supports that claim under our criteria.

\ifdefined\ICLRArxivVersion
\begin{table}[!htbp]
\else
\begin{table}[H]
\fi
\caption{
Comparison-level coding for the 25 published semantic-ablation comparisons.
Claim is coded as Y (explicit claim), N (no corresponding claim), or ? (unclear).
Control support is coded as S (Supported), U (Unsupported), or I (Indeterminate);
a dash indicates that the corresponding claim was not made.
}
\label{tab:app-published-audit}
\centering
\scriptsize
\setlength{\tabcolsep}{3.5pt}
\renewcommand{\arraystretch}{1.02}

\begin{tabular}{@{}llcccc@{}}
\toprule
& &
\multicolumn{2}{c}{Content sensitivity} &
\multicolumn{2}{c}{Predictive utility} \\
\cmidrule(lr){3-4}\cmidrule(lr){5-6}
Study & ID & Claim & Support & Claim & Support \\
\midrule

LIFT       & C01-A & Y & S & Y & U \\
LIFT       & C01-B & Y & S & Y & U \\
LIFT       & C01-C & N & -- & Y & U \\
LIFT       & C01-D & N & -- & Y & I \\

\addlinespace
TabLLM     & C03-A & Y & S & N & -- \\
TabLLM     & C03-B & N & -- & N & -- \\
TabLLM     & C03-C & Y & U & N & -- \\

\addlinespace
PLATO      & C04-A & N & -- & Y & I \\
PLATO      & C04-B & N & -- & Y & U \\
PLATO      & C04-C & N & -- & N & -- \\

\addlinespace
CARTE      & C05-A & N & -- & Y & U \\
CARTE      & C05-B & N & -- & Y & I \\
CARTE      & C05-C & N & -- & Y & U \\

\addlinespace
FeatLLM    & C06-A & N & -- & Y & I \\
TabuLa-8B  & C07-A & ? & -- & Y & S \\

\addlinespace
ConTextTab & C08-A & N & -- & Y & U \\
ConTextTab & C08-B & N & -- & Y & U \\
ConTextTab & C08-C & N & -- & Y & U \\
ConTextTab & C08-D & N & -- & Y & U \\
ConTextTab & C08-E & N & -- & Y & I \\
ConTextTab & C08-F & N & -- & ? & I \\

\addlinespace
TabSTAR    & C09-A & N & -- & ? & S \\
TabSTAR    & C09-B & N & -- & Y & I \\

\addlinespace
TARTE      & C10-A & N & -- & N & -- \\
TARTE      & C10-B & N & -- & Y & U \\

\bottomrule
\end{tabular}
\end{table}

Four comparisons make an explicit content-sensitivity claim.
Three of these use controls that support the claim, while one uses a control that does not.

Eighteen comparisons make an explicit predictive-utility claim.
Of these, 1 uses a control that satisfies all three requirements, 11 use controls that fail at least one requirement, and 6 remain indeterminate because the published description does not establish all required properties.
Five additional comparisons make no predictive-utility claim, and two have unclear claim scope.

These counts describe the 25 comparisons included in this audit and are not intended to estimate the prevalence of unsupported controls in the broader literature.
An unsupported coding means that the reported control does not identify the stated claim under our criteria; it does not imply that the substantive claim is false or that the underlying method is invalid.

Preservation accounts for most of the coding disagreement reported above.
When the published description did not make clear whether other predictive information was preserved, we retained the Indeterminate category rather than assigning Supported or Unsupported.

\subsection{Detailed coding decisions}

Several cases required additional judgment about the scope of the reported claim
or about what information the control preserved.

For LIFT C01-C, we interpret the reported claim as a claim about the predictive
benefit of feature names.
The no-name prompt also removes task descriptions and categorical-value
verbalizations, so the comparison changes more than feature-name information.
We therefore code Preservation as not satisfied for this claim.

For TabLLM C03-B, \emph{List Only Values} removes feature names together with
the explicit key--value structure of the List Template.
Because our feature-name utility comparison requires the remaining input
structure and feature identity to be preserved, this control does not satisfy
Preservation.

For TabLLM C03-C, categorical values are relabeled, but continuous values are
first discretized into ten bins before permutation.
The comparison therefore changes numerical predictive information in addition
to the supplied value semantics.
We consequently do not treat it as isolating content sensitivity to value
semantics alone.

For CARTE C05-A, replacing the string feature initialization with MinHash changes the feature representation, so Preservation is not satisfied.

For FeatLLM C06-A, the evaluated component is the feature-description block supplied for rule generation.
Whether \emph{-Description} also removed the same descriptions from the rule-parsing prompt is not reported, so P4 is Unclear and the control is Indeterminate.

For ConTextTab C08-A--D, the ablations replace the semantic representations of
categorical or string features with ordinal, MinHash, Gap, or AutoGluon
encoders.
These comparisons therefore change the feature representation itself in
addition to removing the evaluated semantic contents, so they do not
satisfy Preservation for a predictive-utility claim.
C08-E replaces column names with generic identifiers, but the published description does not establish whether the same learning procedure was used, so Setup is Unclear and the control is Indeterminate.
The same issue applies to C08-F.

For TabSTAR C09-B, the paper does not state whether removing numerical information from the verbalization also changes the separate numerical input.
Preservation and Setup therefore remain Unclear, and the control is Indeterminate.

\subsection{Source evidence for coded claims}

Claim coding follows the interpretation stated by the original study rather
than the direction or magnitude of the reported ablation result.
For every comparison coded as making an explicit or unclear predictive-utility
claim, Table~\ref{tab:app-published-claim-provenance} records the interpretation
used for coding and the corresponding location in the original paper.
Comparisons for which we found no predictive-utility claim remain coded as
having no such claim.

\ifdefined\ICLRArxivVersion
\begin{table}[!htbp]
\else
\begin{table}[p]
\fi
\caption{
Source evidence used to code predictive-utility claims in the published-ablation audit.
The table includes comparisons with explicit or unclear predictive-utility claims.
Claim basis summarizes the interpretation given in the original study.
}
\label{tab:app-published-claim-provenance}
\centering
\scriptsize
\setlength{\tabcolsep}{3pt}
\renewcommand{\arraystretch}{1.04}

\begin{tabular}{@{}llcP{0.50\linewidth}P{0.20\linewidth}@{}}
\toprule
Study & ID & Claim & Claim basis & Source \\
\midrule

LIFT & C01-A & Y &
The gain over shuffled names is attributed to correct feature--value association in prompt format I. &
Sec.~4.1 / Table~9, p.~8 \\

LIFT & C01-B & Y &
The same interpretation attributes the gain in prompt format II to correct feature--value association. &
Sec.~4.1 / Table~9, p.~8 \\

LIFT & C01-C & Y &
Correctly incorporating feature names is stated to improve performance except on CMC; this interpretation also covers the no-name comparison. &
Sec.~4.1 / Table~9, p.~8 \\

LIFT & C01-D & Y &
The same qualified feature-name benefit is stated for the second no-name comparison. &
Sec.~4.1 / Table~9, p.~8 \\

PLATO & C04-A & Y &
Broader domain knowledge is interpreted as improving prediction beyond a feature-only knowledge graph. &
Sec.~4.1 / Table~3, pp.~7, 9 \\

PLATO & C04-B & Y &
Feature information in the knowledge graph is interpreted as improving prediction relative to the no-KG configuration. &
Sec.~4.1 / Table~3, pp.~7, 9 \\

CARTE & C05-A & Y &
Semantic-similarity encoding is described as important for exploiting external information, especially when the table itself provides little information. &
App.~C.3 / Fig.~10, p.~21 \\

CARTE & C05-B & Y &
Edge information is described as important for capturing table context and achieving strong predictive performance. &
App.~C.3 / Fig.~10, p.~21 \\

CARTE & C05-C & Y &
The attention layer is described as important for capturing table context and achieving strong predictive performance. &
App.~C.3 / Fig.~10, p.~21 \\

FeatLLM & C06-A & Y &
Feature descriptions and reasoning instructions are interpreted as using LLM prior knowledge to improve performance, especially in few-shot settings. &
Sec.~4.2 / Table~4, p.~8 \\

TabuLa-8B & C07-A & Y &
Meaningful headers are explicitly described as benefiting prediction at small shot counts, with the benefit decreasing as more labeled examples are available. &
App.~F.2 / Fig.~12, pp.~26--27 \\

ConTextTab & C08-A & Y &
The ordinal-encoding ablation is interpreted as losing performance because feature semantics are discarded. &
Sec.~5.1 / Table~2, pp.~8--9 \\

ConTextTab & C08-B & Y &
The MinHash ablation is interpreted as reducing the use of feature semantics. &
Sec.~5.1 / Table~2, pp.~8--9 \\

ConTextTab & C08-C & Y &
The same feature-semantics interpretation is applied to the Gap-encoding comparison. &
Sec.~5.1 / Table~2, pp.~8--9 \\

ConTextTab & C08-D & Y &
The same interpretation is applied to the AutoGluon-encoder comparison. &
Sec.~5.1 / Table~2, pp.~8--9 \\

ConTextTab & C08-E & Y &
The loss after replacing column names with generic identifiers is interpreted as evidence that the model benefits from column-name semantics. &
Sec.~5.1 / Table~2, pp.~8--9 \\

ConTextTab & C08-F & ? &
Enriched column descriptions are reported to provide a slight improvement, explicitly qualified as not statistically significant. &
Sec.~5.1 / Table~2, pp.~8--9 \\

TabSTAR & C09-A & ? &
Adding quantile information beyond bins is described as having limited impact despite a small average improvement. &
App.~G.4, p.~52 \\

TabSTAR & C09-B & Y &
Including numerical information in verbalization is interpreted as improving performance relative to feature names alone. &
Sec.~6 (Q3) / Table~4, pp.~9--10; App.~G.4, p.~52 \\

TARTE & C10-B & Y &
The MinHash comparison is interpreted as showing the importance of FastText semantic similarity for prediction. &
Sec.~4.3 / Fig.~6, pp.~9--10 (author version) \\

\bottomrule
\end{tabular}
\end{table}

\ifdefined\ICLRArxivVersion
\let\AppendixSavedFloatBarrier\FloatBarrier
\let\FloatBarrier\relax
\fi
\section{Limitations and Future Work}
\ifdefined\ICLRArxivVersion
\let\FloatBarrier\AppendixSavedFloatBarrier
\fi
\label{sec:app-limitations}

Our framework requires a reference that removes the semantic component being evaluated while preserving the rest of the predictive comparison.
Such a reference is not always available.
In some settings, removing semantic knowledge can also change other predictive information or the model input, while in others the available description may not provide enough detail to verify what is preserved.
In these cases, we treat predictive utility as not separately evaluable rather than infer it from a control that changes additional factors.

Our experiments focus on settings in which the evaluated semantic component can be identified from feature names, dataset documentation, codebooks, known schema structure, or explicit relations.
The framework is less straightforward when the relevant information is distributed across opaque or high-dimensional representations and cannot be isolated as a well-defined component.
Future work could study how to construct and validate suitable references in such settings.

Predictive utility can also vary with the chosen reference and with how easily the learner can recover the removed information from other inputs or labeled target examples.
Our experiments examine this dependence across several learners and reference constructions, but they do not cover all possible models or representations of semantic knowledge.
Broader evaluations could test whether the same patterns hold for other tabular foundation models and cross-schema learning systems.

Finally, our published-ablation analysis is a bounded audit rather than a systematic review of the tabular-learning literature.
Its purpose is to examine whether specific claim--control pairs satisfy the proposed criteria, not to estimate how common particular evaluation practices are.
A larger systematic review could examine how often these patterns arise across a broader range of methods and domains.

\section{Reproducibility and Artifacts}
\label{sec:app-extraction}

\subsection{Software environment and data-access boundaries}

\begin{table}[h]
\caption{
Software versions used across the principal analyses and model-specific extensions.
Model-specific environments are listed separately where they differ from the principal analysis environment.
}\label{tab:app-software-environment}
\centering\small
\begin{tabular}{ll}
\toprule
Component & Version \\
\midrule
Python (principal analyses) & 3.10.19 \\
NumPy & 2.2.6 \\
pandas & 2.3.3 \\
scikit-learn & 1.7.2 \\
XGBoost & 3.1.3 \\
TabPFN & 7.0.0 \\
PyTorch & \texttt{2.12.0.dev20260408+cu128} \\
Transformers (documentation extraction) & 5.12.0 \\
Python (TabLLM / learner extensions) & 3.10.20 \\
Transformers (TabLLM / TransTab) & 4.30.0 \\
Datasets (TabLLM) & 2.14.7 \\
TransTab (commit) & \texttt{fdb34cf3} \\
CARTE & \texttt{carte-ai==0.0.26} \\
Python (examination support/reference, value-redundancy, CARTE) & 3.10.20 \\
Transformers (original TransTab run) & 5.12.0 \\
Relation-capacity sensitivity: Python / NumPy / XGBoost
& 3.8.8 / 1.24.4 / 2.1.4 \\
Null-source extension: NumPy / scikit-learn / XGBoost
& 1.24.4 / 1.3.2 / 2.1.4 \\
\bottomrule
\end{tabular}
\end{table}

The supplied one-thread replication reproduces all 400 comparable stored
HistGradientBoosting nMSE values exactly. This establishes equality of the
recorded numerical outputs; cross-thread bitwise identity is not claimed.
A second planned HRS measurement--correspondence interaction experiment was not run because it requires new authorization for licensed HRS microdata.

\subsection{Supplementary artifact map}

\begin{table}[H]
\caption{Map of reproducibility artifacts available in the public repository.}
\label{tab:app-artifact-map}
\centering\scriptsize
\setlength{\tabcolsep}{4pt}
\begin{tabular}{P{0.30\linewidth}P{0.62\linewidth}}
\toprule
Artifact directory & Contents \\
\midrule
\path{provenance/} &
complete record of when each analysis was designed and the associated decisions \\
\path{tabllm/} &
complete frozen reference library, reference-construction records, reference-sensitivity results, minimum and maximum utility for each dataset, results for all references with 512 labeled adaptation examples, and utility changes with labeled examples for each reference \\
\path{robustness/measurement/} &
absolute AUROC results for the focused measurement analysis \\
\path{robustness/correspondence/} &
correspondence results and record of the clean rerun \\
\path{robustness/relation/} &
evidence supporting the retained relations, results after removing inputs used to compute relation values, and checks of how well relation values can be recovered \\
\path{robustness/interaction/} &
complete interaction results, endpoint robustness checks, label corruption checks, and source-data integrity checks \\
\path{robustness/resolution/} &
minimum detectable effect calculations and results for each endpoint or basin \\
\path{model_coverage/} &
complete TransTab and CARTE results \\
\path{method_contract/} &
complete prompt components, rendered prompts, schemas, acceptance rules, raw responses, and generation manifests \\
\path{extraction/} &
complete seven-model measurement-code extraction panel \\
\path{published_audit/} &
initial human coding, independent P1--P6 checks, final consensus,
source evidence, and adjudication history \\
\path{reproducibility/} &
exact software and model versions, environment details, and checks across thread counts \\
\bottomrule
\end{tabular}
\end{table}

\FloatBarrier

\ICLRFinishAppendix

\fi

\end{document}